\documentclass{article}

\usepackage{arxiv}

\usepackage[utf8]{inputenc} 
\usepackage[T1]{fontenc}    
\usepackage{hyperref}       
\usepackage{url}            
\usepackage{booktabs}       
\usepackage{amsfonts}       
\usepackage{nicefrac}       
\usepackage{microtype}      
\usepackage{lipsum}

\usepackage{graphicx}
\usepackage{amssymb, amsmath}
\usepackage{lineno}
\usepackage{color}
\usepackage{multirow}
\usepackage{subcaption}
\usepackage{float}
\usepackage{algorithm2e}
\usepackage{xcolor}
\usepackage{array}
\usepackage[numbers]{natbib}
\usepackage{longtable}

\makeatletter
\newcommand{\thickhline}{%
    \noalign {\ifnum 0=`}\fi \hrule height 1pt
    \futurelet \reserved@a \@xhline
}
\newcolumntype{"}{@{\hskip\tabcolsep\vrule width 1pt\hskip\tabcolsep}}
\makeatother

\title{An interpretable semi-supervised classifier using two different strategies for amended self-labeling}

\author{
  Isel Grau\thanks{This work was supported by the IMAGica project, financed by the Interdisciplinary Research Programs and Platforms (IRP) funds of the Vrije Universiteit Brussel; and the BRIGHTanalysis project, funded by the European Regional Development Fund (ERDF) and the Brussels-Capital Region as part of the 2014-2020 operational program through the F11-08 project ICITY-RDI.BRU (icity.brussels).} \\
  Artificial Intelligence Lab \\
  Vrije Universiteit Brussel, Belgium \\
  \texttt{igraugar@vub.be} \\
  \And
  Dipankar Sengupta \\
  Centre for Cancer Research and Cell Biology \\
  Queens University Belfast, United Kingdom \\
  \And
  Maria Matilde Garc\'ia Lorenzo \\
  Department of Computer Science \\
  Universidad Central ``Marta Abreu" de Las Villas, Cuba \\
  \And
  Ann Now\'e \\
  Artificial Intelligence Lab \\
  Vrije Universiteit Brussel, Belgium \\
}

\begin{document}
\maketitle

\begin{abstract}
In the context of some machine learning applications, obtaining data instances is a relatively easy process but labeling them could become quite expensive or tedious. Such scenarios lead to datasets with few labeled instances and a larger number of unlabeled ones. Semi-supervised classification techniques combine labeled and unlabeled data during the learning phase in order to increase classifier's generalization capability. Regrettably, most successful semi-supervised classifiers do not allow explaining their outcome, thus behaving like black boxes. However, there is an increasing number of problem domains in which experts demand a clear understanding of the decision process. In this paper, we report on an extended experimental study presenting an interpretable self-labeling grey-box classifier that uses a black box to estimate the missing class labels and a white box to explain the final predictions. Two different approaches for amending the self-labeling process are explored: a first one based on the confidence of the black box and the latter one based on measures from Rough Set Theory. The results of the extended experimental study support the interpretability by means of transparency and simplicity of our classifier, while attaining superior prediction rates when compared with state-of-the-art self-labeling classifiers reported in the literature.
\end{abstract}


\section{Introduction}
\label{sec:intro}

Gathering data examples for training a machine learning classifier in a real-world scenario is often simple, but the process of assigning labels to the examples can be costly in terms of money, time or effort. In such scenarios we might obtain datasets with more unlabeled than labeled data. This is often the case in applications such as image classification \cite{gong2016multi}, industrial fault classification \cite{YIN201988}, sentiment analysis \cite{XU2019120}, speaker identification \cite{fazakis2015speaker} and bioinformatics or medical applications \cite{XIE2019237}. Semi-supervised classification (SSC) techniques arise from the need to address this problem using both labeled and unlabeled data for training a classifier. The aim is to increase the generalization ability of the classifier when compared to a supervised classifier that only uses the available labeled data.

The SSC literature reports several techniques including transductive Support Vector Machines \cite{Bennett1999}, Graph-based methods \cite{blum2001learning}, Generative Mixture Models \cite{fujino2008semisupervised}, Self-labeling techniques \cite{triguero2015self} and more recently semi-supervised Generative Adversarial Networks \cite{salimans2016improved}. In general, state-of-the-art SSC methods involve three main shortcomings that may vary from a specific family of algorithms to the whole field. The first potential issue affecting all SSC models refers to the assumption that the unlabeled data helps elucidating the distribution of the labeled instances. When this assumption is not met in any of its forms, semi-supervised learning may not be useful. Secondly, some techniques such as Graph-based methods mainly focus on transductive learning, i.e. predicting the label for a given set of unlabeled data rather than finding a model capable of predicting the classification of unseen instances with a proper generalization. Thirdly, while self-labeling approaches such as Co-training \cite{hady2008co}, Self-training \cite{yarowsky1995unsupervised} and their variants perform quite well in terms of accuracy, they often result in complex structures combining several classifiers and failing to give the user insight in how the classification process comes about.

An increasing requirement observed in machine learning is to obtain not only precise models but also interpretable ones. End users often demand an insight into how an algorithm arrives at a particular outcome and need an explanation of the decisions to some extent. In general, explainable artificial intelligence is starting to be a central concern in both governing and research communities. For example, the EU General Data Protection Regulation includes a right to obtain an explanation on the decisions made by an algorithm affecting human beings \cite{goodman2017european}. This regulation might limit the potential of using artificial intelligence in a variety of domains, unless we start developing more transparent models.

Recent studies \cite{doshi2017towards,gilpin2018explaining,lipton2016mythos,BARREDOARRIETA202082} formalize terms such as interpretability or explainability in sometimes overlapping concepts. However, a common conclusion is that a certain grade of global interpretability can be reached through the use of more transparent techniques as proxies for solving a task. In this paper, we refer to intrinsically interpretable models (e.g., linear regression, decision trees or rule induction algorithms) as white boxes, as opposed to the less interpretable black-box ones (e.g. artificial neural networks or support vector machines). Black boxes are normally more accurate techniques that learn exclusively from data but they are not easily understandable at a global level. Whereas white boxes refer to models which are constructed based on laws or principles of the problem domain, or those who are built from data but their structure allows for  explanations or interpretation, since pure white boxes rarely exists \cite{nelles2013nonlinear}. Intrinsically interpretable models can be recommended when a transparent model that can be inspected as a whole is needed and the prediction problem does not require a very powerful technique. On the other hand, agnostic post-hoc methods \cite{ribeiro2016,NIPS2017_7062} are a suitable alternative when a black-box is already built and we need to compute explanations for input and output pairs, preserving accuracy. However, post-hoc methods generate explanations that are often local or limited to feature attribution rather than a holistic view of the model. Grey-box models, i.e. using white boxes as surrogates for distilling previously trained black boxes are an approach in between intrinsically interpretability and model agnostic post-hoc. While the white boxes attempt to explain the problem domain directly, the grey-boxes are devoted to explain the domain by approximating the predictions produced by a black-box classifier.

In this paper, we study the SSC problem from the interpretability angle. We conduct a detailed revision of methods reported in the literature and discuss their shortcomings when interpretability comes to play. We explore the performance of our semi-supervised classifier termed \emph{self-labeling grey-box} (SlGb) \cite{grau2016,grau2018}, which exploits the strength of black-box models being good classifiers with the interpretability of white boxes. In terms of interpretability, we refer to a grey-box model as the combination of a black-box model with a white-box one. Our classifier uses a black box to estimate the decision class for unlabeled instances in order to increase the amount of training data. Afterwards, our approach builds a surrogate white-box classifier from the enlarged dataset that allows explaining the predictions. In addition we explore the effects of using two weighting strategies to reduce the effect of misclassifications when building the enlarged dataset. The former is based on the black box's confidence for the inferred class label, while the latter is based on granular computing principles. The use of an enlarged dataset combined with a weighting strategy results in a white box with improved prediction rates. Numerical experiments using 55 datasets in different settings show that our proposal attains a good balance between prediction rates and explainability, while outperforming most state-of-the-art methods.

The rest of this paper is structured as follows. Section \ref{sec:related-work} provides an overview of state-of-the-art SSC algorithms reported in the literature and their interpretability, while making emphasis on self-labeling techniques. Section \ref{sec:model} describes the SlGb approach and Section \ref{sec:amend} depicts two alternatives for the amending of the self-labeling performed by the black-box classifier. Section \ref{sec:exp} introduces the numerical simulations and an extensive discussion covering the performance and interpretability of the SlGb. Section \ref{sec:conclusions} formalizes the concluding remarks and research directions to be explored in the future.

\section{Semi-supervised Classification Methods and their Interpretability}
\label{sec:related-work}

In supervised classification the goal is to identify the right category (among those in a predefined set) to which an observation belongs. These observations (henceforth called instances) are often described by a set of numerical and/or nominal attributes. Solving this problem implies to define a mapping $f: X \rightarrow Y$ that assigns to each instance $x \in X$, described by a set of attributes $A =\{a_1\,\ldots,a_p\}$, a decision class $y \in Y$. The mapping is learned from data in a supervised fashion, i.e., by relying on a set of previously labeled examples, used to train the classifier.

Semi-supervised techniques attempt to use both labeled and unlabeled instances during the learning process for increasing the prediction capacity when only labeled data is used. More formally, in a SSC scenario we have a set of $m$ instances $L=\{l_1,\dots,l_m\}$ which are associated with their respective class labels in $Y$, and a set of $n$ unlabeled instances $U=\{u_1,\dots,u_n\}$, where usually $n>m$. In the context of SSC, the classifier performance can be evaluated in two settings: (1) transductive learning, which only attempts to predict the labels for the given unlabeled instances in $U$; or (2) inductive learning, which tries to infer a mapping $g: L \cup U \rightarrow Y$ for predicting the class label of any instance associated with the classification problem. 

In this section, we review the main state-of-the-art methods for semi-supervised classification, including an analysis of their interpretability. Here, we evaluate the interpretability as the inherent model transparency, as described in \cite{lipton2016mythos}.

\subsection{Semi-supervised Classification Methods}
\label{sec:related-work:families}

As mentioned, SSC methods often involve assumptions about the distribution or characteristics of the unlabeled data \cite{zhu05survey}. For example, transductive Support Vector Machines (tSVMs) \cite{Bennett1999} assume that the decision boundary lies in a low-density region. This method uses unlabeled data for maximizing the margin between the different classes by placing the decision boundaries in sparse regions. However, given the fact that the complexity of the optimization problem increases in the semi-supervised setting, its computational burden is quite high and it does not scale well for large-scale data. Recent studies \cite{Cevikalp2017,Li2018} try to overcome this limitation by using the concave-convex procedure and variations of stochastic gradient descent to solve the optimization problem. Although SVMs are a powerful technique with a strong mathematical framework for building classifiers, it has the drawback of working as a black box from the interpretability point of view. The lack of transparency of SVMs does not allow them to produce explanations or interpretations of the obtained model. In this case, the use of post-hoc methods for generating explanations is necessary when requiring explanations over the obtained predictions.

Graph-based methods \cite{blum2001learning} assume that high-dimensional data lie on a low-dimensional manifold \cite{Chapelle2010}. These methods represent the data space as a graph (i.e., if two instances are strongly connected, then they likely belong to the same class) and estimate a continuous function which is close enough to the label values, with the ultimate goal of propagating labels between similar instances. Recent works on Label Propagation methods \cite{Gong2017,Fan2018} are mainly focused on the construction of an effective graph over data with complex distribution and reducing the risk of error propagation through outliers. This approach could be interpretable to some extent by visually inspecting the obtained graph from the structural point of view, allowing some transparency at the parameters level. A first work toward this direction can be found in \cite{rustamov2018interpretable}, where the authors propose a flow sub-graph framework which visualizes the path along the information flow from a source labeled instance to a target unlabeled instance. These sub-graphs can be seen as rather local explanations in the form of visualizations of the model. Their usability is limited to data that can be represented in the graph replacing the abstract representation of the node (e.g. images). A more general option is to obtain kNN-like explanations with examples by leveraging the graph structure, e.g. ``the predicted label of instance $x_i$ was propagated from instances $x_1$, $x_2$ and $x_3$''.

A third approach assumes that the data follow an identifiable mixture distribution (Generative Mixture Models \cite{fujino2008semisupervised}), henceforth they learn a joint probability for identifying the mixture components using the unlabeled data. This approach may be convenient when the available data produce well-separated clusters \cite{zhu05survey}, but most of the time the joint distribution is not easily identifiable. Here the estimated mixture distribution could be interpretable at a very high abstraction level if the representation space of the problem at hand is not too complex. However, the unlabeled data could have a negative effect on algorithm's performance if the generative model is wrong. From the interpretability point of view, the classification of a new instance can leverage the Bayes rule for building a (rather abstract) explanation: ``$y_j$ is the most probable value of $y$ for $x_i$ since the probability $p(x_i)$ is high when $y=y_i$''. Moreover, the estimated mixture distribution could only be visualized in a low-dimensional feature space for gaining insights into the clusters found by the model. In our opinion, GMMs require the use of post-hoc methods or global surrogates for gaining in interpretability of their results. An interesting work in this direction includes generating rectangular regions from the clusters and transforming them into rules \cite{interpretableGMM}. 

More recently, deep architectures have been explored by extending graph-based methods \cite{weston2012deep} and generative models \cite{kingma2014semi}. Particularly successful has been the extension of Generative Adversarial Networks (GAN) \cite{goodfellow2014generative} to the SSC context \cite{odena2016semi,salimans2016improved}. For example, Feature Matching GANs \cite{salimans2016improved} use a discriminator for $c+1$ labels instead of the binary ``real/fake" distinction, where the first $c$ are the class labels of the problem and $c+1$ corresponds to the generated instances. The authors in \cite{nips2017} theoretically analyze whether a good generator and a good discriminator for semi-supervised learning can be obtained at the same time. The study concludes that the generator should be ``bad'' in the sense of assigning high probabilities to low-density regions of the input space according to the true distribution, in order to complement the true data distribution and improve the semi-supervised performance. Regarding interpretability, deep neural networks are black-box models that need post-hoc procedures for generating explanations of their predictions. The majority of contributions are focused on local surrogate models or feature importance methods specially designed for deep multilayer, convolutional or recurrent neural networks \cite{Chakraborty2017,van2020survey}. Interesting works connected to the semi-supervised setting include learning disentangled latent representations in a variational autoencoder, i.e. latent variables with an interpretable meaning coming from labeled data are added to the latent representation \cite{NIPS2017_7174}. These latent interpretable variables can be used later on for inspecting their influence in the prediction.

Finally, self-labeling refers to a wide family of very powerful and versatile wrapper methods that employ one or more base classifiers for enlarging the available labeled dataset assuming the predictions they produce on the unlabeled data are correct. Since our contribution falls within this category, we decided to revise those SSC methods in a separate subsection to gain in clarity.

\subsection{Self-labeling Techniques}
\label{sec:related-work:selflabeling}

According to \cite{triguero2015self}, self-labeling techniques can be categorized into single-view or multi-view methods based on whether they need one or multiple datasets for learning. Self-training approaches \cite{yarowsky1995unsupervised} are single-view wrapper classifiers, which rely on the prediction of only one base classifier to repeatedly increase the size of the labeled dataset by predicting the unlabeled instances. The instances are added incrementally, in batch \cite{halder2010ant} or in an amending procedure \cite{li2005setred}. The use of amending procedures allows selecting or weighting the self-labeled instances for enlarging the labeled dataset, hence avoiding error propagation.

The multi-view methods assume that the data space can be described from two or more different viewpoints. These different views normally correspond to distinct sets of attributes describing the same instances \cite{Witten2017467}. A classic example of multi-view methods is the Co-training \cite{blum1998combining} approach, where different classifiers are trained separately, each using a different attribute subset. Thereafter, the prediction of each classifier over the unlabeled dataset is used for enlarging the training set of the other. Other alternatives using multiple classifiers but not needing multi-view datasets are Democratic Co-learning \cite{zhou2004democratic}, Tri-training \cite{zhou2005tri}, Co-training by committee \cite{hady2008co} and Co-Forest \cite{li2007improve} which use several base classifiers of the same type. Tri-training uses three base classifiers that collaborate in the learning process by labeling an unlabeled example if the other two classifiers agree. An alternative to Co-training is Co-training by committee, which does not require multi-view nor different learning algorithms, and explores different ensemble strategies with Bagging as the best performing one. Similarly, Co-Forest adopts a Random Forest classifier as an alternative for Co-training.

Self-labeling techniques are easy to implement and apply to almost all existing classifiers \cite{fazakis2016self,albinati2015ant,karlos2016locally,vluymans2016fuzzy}. A wide experiment conducted in \cite{triguero2015self} shows that CoTraining using Support Vector Machines as a base classifier \cite{hady2008co}, TriTraining using C4.5 decision tree \cite{zhou2005tri}, CoBagging using C4.5 \cite{hady2008co} and Democratic Co-learning (as an ensemble of Na\"ive Bayes, C4.5 and K-Nearest Neighbors) \cite{zhou2004democratic}, are the best performing self-labeling classifiers evaluated against a comprehensive collection of benchmark datasets. Other semi-supervised classifiers that have demonstrated competitive performance in a variety of datasets are self-training using logistic model trees \cite{fazakis2016self}, differential evolution \cite{Wu2018} or naive Bayes \cite{karlos2016locally}. 

In terms of interpretability, a self-training scheme producing a simulatable model (e.g., relatively simple tree structure) as the final classifier can be considered a transparent model. More complex schema such as Tri-training, Co-Bagging, Co-Forest or Co-training are less likely to be interpretable due to the collaborative nature of the algorithms and the complexity of the resulting structure. However, the ensemble character of self-labeling is a perfect match with the use of local or global surrogate models for explaining predictions. Combining base classifiers using self-labeling in a way that the resulting ensemble works as a surrogate white box is the challenge we want to address. In the next section, we describe a simple yet effective self-labeling method which uses two base classifiers, a black box and a white box, for reaching a suitable trade-off between performance and interpretability.

\section{Self-labeling Grey-box Approach}
\label{sec:model}

In this section we describe the \textit{self-labeling grey-box} proposed in our previous works \cite{grau2016}\footnote{The authors in \cite{pintelas2020grey} report on an ensemble grey-box strongly inspired in our previous work \cite{grau2016}. They claim that their main contribution is the use of the confidence of the black box predictions as amending of the self-labeling, however this idea was earlier proposed in our previous work \cite{grau2018}. The only difference between their method and our model is that they use an iterative process to select the instances to be included in the enlarged dataset. However, this seems to be a strange approach considering that the confidence amending reduces the need of such an iterative process.}. Here, we use a black-box classifier to predict the decision class of the unlabeled instances, while a surrogate white box is used to build an interpretable predictive model (e.g., a rule-based approach), based on the whole instance set. The aim is to outperform the base white-box component using only the originally labeled data, while maintaining a good balance between performance and interpretability. It is worth mentioning that the main motivation behind SlGb is not to outperform the most complicated state-of-the-art algorithms but to provide a simple approach allowing for interpretability. In other words, we should be able to produce competitive solutions without significantly increasing the complexity inherent to the base classifiers.

The learning process is performed in a sequential order. In a first step, we provide the available labeled dataset $(L,Y)$ to a black-box classifier for training. Once the supervised learning is completed, the black-box component has learned a function $f:L \rightarrow Y$, where $f \in F$, being $F$ the hypothesis space that associates each instance with a class label. The $f$ function can be computed from the scoring function $h:L \times Y \rightarrow [0,1]$ such that $f(x)= argmax_{y\in Y} \{h(l,y)\}, l \in L$. Thereafter, the trained black-box component is used for generating new tuples $(u,y)$ by mapping all unlabeled instances $u \in U$ to a class label $y \in Y$ as $y=f(u)$, adding a self-labeling character to the approach. From this step we obtain an enlarged training set $(L \cup U,Y)$ comprising the original labeled instances and the extra labeled ones.

\begin{figure}[!ht]
  \centering
  \includegraphics[height=5cm]{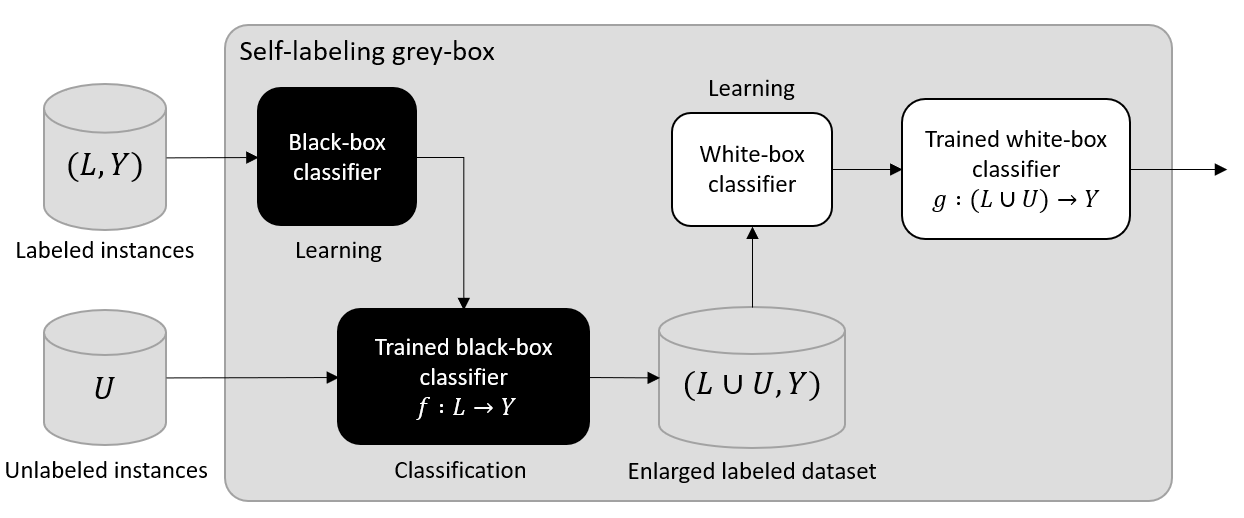}
  \caption{Blueprint of the SlGb architecture. During the first step, labeled data is used for training a black-box model, which assigns labels to the unlabeled data. Later on, a white-box surrogate model is trained on the enlarged dataset, thus resulting in an interpretable model.}
  \label{fig:grey-model1}
\end{figure}

In the second step, the enlarged training set $(L \cup U,Y)$ is used to train a surrogate white-box classifier. Once the learning process in the white-box component is completed, we obtain a function $g:(L \cup U) \rightarrow Y$ resulting in a classifier which is more likely to have better generalization capabilities than the original white-box component, when trained on only the labeled data. Figure \ref{fig:grey-model1} summarizes this process.

When applying self-labeling, we should be aware of the risk of having imbalanced data with respect to the class labels. It might be easier to obtain unlabeled data of a certain class, for example, in the context of credit fraud detection or rare diseases classification. In order to deal with this problem, our approach additionally incorporates a simple strategy for balancing instances as a preprocessing step. This weight is computed as: 

\begin{equation}
\label{eq:weightslab}
    w_{(l_j,y_i)} = {|L_{[y_{min}]}|}/{|L_{[y_i]}|}
\end{equation}

\noindent where $L_{[y_i]}, L_{[y_{min}]} \subset L$ denote the sets of labeled instances that are mapped to the class label $y_i$ and the minority class $y_{min}$, respectively. In this way we assign higher importance to instances belonging to the minority class.

In general, the SlGb approach is only based on the general assumption of SSC methods: the distribution of unlabeled instances helps elucidate the distribution of all examples. In addition, our approach allows retaining the inherent interpretability of the chosen white-box surrogate. According to the taxonomy proposed in \cite{triguero2015self}, our approach can be categorized as follows:

\begin{itemize}
  \item[]\textbf{single-view:} the SlGb classifier does not need different attribute sets for describing the instances, adding simplicity to the model;
  \item[]\textbf{multi-classifier:} two different base classifiers are used, connected in a sequential process, the first classifier should be a good performing black-box supervised classifier, whereas the second should be a white-box technique guaranteeing interpretability to the final model;
  \item[]\textbf{multi-learning:} the learning process comprises two steps, where two different learning algorithms are used depending on the base classifiers.
\end{itemize}

It can be noticed that the performance of the whole SlGb approach largely depends on the prediction capability of the black-box classifier when classifying unseen instances. Obviously, like any other machine learning algorithm, when solving application problems the performance will also depend on the quality of the data and the application of domain-dependent preprocessing steps \cite{lazar2012batch}. However, in the context of self-labeling, the classification mistakes can reinforce themselves if no amending procedure is used during self-training. Therefore, in the next section we describe two amending strategies for the self-labeled instances, in order to prevent the error from propagating through the model. 

\section{Amending Strategies}
\label{sec:amend}

In this section, we describe two strategies for weighting the instances that result from the self-training stage. The goal is to improve the quality of the final model either in terms of performance or interpretability. The first strategy uses the confidence of the predictions made by the base black box and the second one focuses on the possible inconsistency of the enlarged dataset. Therefore, both procedures assign more importance to more reliable instances in the second learning step, avoiding the propagation of errors or inconsistent information. 

\subsection{Using the Class Membership Probabilities of the Black-box Classifier}
\label{sec:amend:prob}

For the first strategy, the amending process for each unlabeled instance $u$ is based on the class membership probability, which is computed by the black-box classifier in the self-labeling. The weights are assigned to the instances after they are labeled by the black-box classifier, thus expressing the confidence degree associated to the self-labeling process. Equation \eqref{eq:weightsunlab} shows how to compute the weight $w_{(u_k,y_i)}$ using the scoring function of the black-box base classifier $h(u_k,y_i)$ that expresses the class membership probability of $u_k$ being correctly assigned to the $y_i$ class,

\begin{equation}
\label{eq:weightsunlab}
  w_{(u_k,y_i)} = h(u_k,y_i) .
\end{equation}

The proposed amending strategy constitutes an alternative to the use of incremental or batch procedures. Our amending does not need several iterations, thus reducing the computational burden of the self-labeling process. The pseudo-code in Algorithm \ref{alg} formalizes the method and incorporates the amending step in the general scheme.

\begin{algorithm}[!hb]
\caption{SlGb learning algorithm with confidence amending. 
\label{alg}}
\DontPrintSemicolon
\KwData{Labeled instances $(L,Y)$, Unlabeled instances $U$}
\KwResult{$g:(L \cup U) \rightarrow Y$}
\Begin{

\tcc{Preprocessing: Weight labeled instances according to Eq. \eqref{eq:weightslab}}
\ForAll {$(l_j,y_i) \in (L,Y)$}{
 	$w_{(l_j,y_i)} \longleftarrow {|L_{min}|}/{|L_i|}$ \;
}

\tcc{Train black-box component with weighted labeled data}
$f,h \longleftarrow blackboxClassifier.fit(L,Y,w)$ \;

\tcc{Self-labeling process: Assign a label to unlabeled instances using black-box inference}
\ForAll{$u_k \in U$}{
 	$y_i \longleftarrow f(u_k)$ \;
    \tcc{Compute weight of instance $u_k$ according to Eq.\eqref{eq:weightsunlab}} 
    $w_{(u_k,y_i)} \longleftarrow h(u_k,y_i)$ \;
    \tcc{Add the instance to enlarge dataset} 
    $(L \cup U,Y) \cup \{(u_k,y_i)\}$\; 
}

\tcc{Train white-box component with the weighted $(L \cup U,Y)$ dataset}
$g \longleftarrow whiteboxClassifier.fit(L \cup U, Y, w)$\;
\Return{$g$}
}
\end{algorithm}

It is important to mention that the black-box classifier should be able to measure calibrated probabilities in order to correctly interpret them as the confidence of its predictions. Not all machine learning models are able to provide probabilities that match with the expected distribution of probabilities for each class. According to a study on different supervised classifiers regarding probabilities estimation \cite{niculescu2005predicting}, logistic regression, multilayer perceptrons and bagged trees naturally provide well calibrated probabilities, whereas others such as boosted trees and SVM produce distorted ones. When the calibration of probabilities is needed, two main options are available: Platt's scaling \cite{platt1999probabilistic} and isotonic regression \cite{zadrozny2001obtaining}. Platt's scaling is more recommended when the distortion in the predicted probabilities has a sigmoid shape, whereas isotonic regression is able to correct any monotonic distortion but it requires large amounts of data for avoiding overfitting.

The amending based on class membership probabilities assumes the ground truth labels are correct and induces the white box to focus its learning on instances that are certain according to that. However, when dealing with limited labeled data we should not discard the existence of noise in the class labels. This can generate class inconsistency, especially when unlabeled data is added from different sources.

\subsection{Using the Inclusion Degree Measures from Rough Set Theory}
\label{sec:amend:incl}

In this subsection we describe a second strategy for amending the enlarged dataset, which is based on the knowledge structures attached to Rough Set Theory \cite{Pawlak1982}. This formalism allows handling uncertainty in the form of inconsistency through the computation of the lower and upper approximations for any set of instances in the decision space. The rough regions associated to these approximations can be used to weight the instances after performing the self-labeling process. Particularly, we assign higher weights to more confident instances as they have more chance to be correctly classified by the base black box.

\subsubsection{Rough Set Theory}
\label{sec:amend:incl:rst}

\emph{Rough Set Theory} (RST) \cite{Pawlak1982} is a mathematical formalism for handling uncertainty in the form of inconsistency. Given a decision system $DS=(\mathcal{U},A\cup \{d\})$ where the universe of instances $\mathcal{U}$ is described by a non-empty finite set of attributes $A$ and its respective decision class $d$, any concept (subset of instances) $X \in \mathcal{U}$ can be approximated by two crisp sets. These sets are called lower and upper approximations of $X$ ($\underline{B} X$ and $\overline{B} X$, respectively) and can be computed taking into account an equivalence relation, as follows:

\begin{equation}
\label{eq:rst:low}
\underline{B} X = \{x \in \mathcal{U}~|~[x]_{B} \subseteq X \}
\end{equation}

\begin{equation}
\label{eq:rst:up}
\overline{B} X = \{x \in \mathcal{U}~|~[x]_{B} \cap X \neq \emptyset \}
\end{equation}

The equivalence class $[x]_{B}$ gathers the instances in the universe $\mathcal{U}$ which are inseparable according to a subset of attributes $B \subseteq A$. From the formulations of upper and lower approximation, we can derive the positive, negative and boundary regions of any subset $X \in \mathcal{U}$. The positive region $\mathcal{P}(X) = \underline{B} X$ includes those instances that are surely contained in $X$; the negative region $\mathcal{N}(X) = \mathcal{U} - \overline{B} X$ denotes those instances that are surely not contained in $X$, while the boundary region $\mathcal{B}(X) = \overline{B} X - \underline{B} X$ captures the instances whose membership to the set $X$ is uncertain, i.e., they might be members of $X$.

The classic RST is regularly defined over a subset of discrete attributes, thus generating a partition of $\mathcal{U}$. A more relaxed formulation of RST establishes the inseparability between instances based on a weak binary relation. Equation \eqref{eq:rst:sim} formalizes the similarity relation used in this paper, which define whether any pair of instances $x_{i}$ and $x_{j}$ can be considered similar,

\begin{equation}
\label{eq:rst:sim}
\mathcal{R}: x_{i} \mathcal{R} x_{j} \rightarrow \delta(x_{i},x_{j}) \geq \varepsilon
\end{equation}

\noindent where $\delta(x_{i},x_{j})$ computes the extent to which $x_{i}$ and $x_{j}$ are deemed inseparable as indicated by the similarity threshold $\varepsilon$. Under this assumption, the universe is arranged in similarity classes that are not longer disjoint but overlapped. In this paper, $\varepsilon = 0.98$ and the inseparability relation is defined as the complement of a distance function, such as the Heterogeneous Euclidean-Overlap Metric \cite{wilson1997improved}. This distance function computes the normalized Euclidean distance between numerical attributes and an overlap metric for nominal attributes. Equations \eqref{eq:rst:heom1} and \eqref{eq:rst:heom2} define this dissimilarity function,

\begin{equation}
\label{eq:rst:heom1}
\delta(x_{i},x_{j}) = \sqrt{\frac{\sum_{t=1}^{|B|}\omega_t \rho_t(x_{i},x_{j})}{\sum_{t=1}^{|B|} \omega_t}}
\end{equation}

\noindent with, 

\begin{equation}
\label{eq:rst:heom2}
\rho_t(x_{i},x_{j}) = \begin{cases}
    0 & \text{if } b_t \text{ is nominal } \wedge x_{i}(t) = x_{j}(t) \\
    1 & \text{if } b_t \text{ is nominal } \wedge x_{i}(t) \neq x_{j}(t) \\
    (x_{i}(t)-x_{j}(t))^2 & \text{if } b_t \text{ is numerical} \\
\end{cases}
\end{equation}

\noindent where $x_{i}(t)$ and $x_{j}(t)$ denote the normalized values of the $t$-th attribute for heterogeneous instances $x_{i}$ and $x_{j}$, respectively, and $\omega_t$ is the information gain of the $b_t$ attribute.

Once the covering of the decision space is generated according to the similarity function, several RST based measures can be computed for measuring the uncertainty contained in a dataset \cite{bello2012rough}. In the following subsection, we adopt one of these measures to weight the instances belonging to the enlarged training set obtained after performing the self-labeling process.

\subsubsection{Inclusion Degree}
\label{sec:amend:incl:edit}

The use of this amending strategy is based on the fact that the black box could produce wrong labels for unlabeled instances. In addition, there is no guarantee that the knowledge concerning the original labeled instances is confident. To address both situations together, we propose a mechanism to weight the instances after the self-labeling process. Unlike the confidence-based strategy, this amending procedure is adopted for the entire enlarged dataset, instead of only the self-labeled instances. Therefore, it treats the uncertainty in the form of inconsistency of the labeled and unlabeled instances together. 

More explicitly, the second weighting strategy is based on the inclusion degree of both labeled and self-labeled instances into the RST granules, thus let $X = L \cup U$ and $d=y$. Let $\mu_{\mathcal{P}_{(y_i)}}^\mathcal{R}(x)$, $\mu_{\mathcal{B}_{(y_i)}}^\mathcal{R}(x)$ and $\mu_{\mathcal{N}_{(y_i)}}^\mathcal{R}(x)$ denote the membership degrees of any instance $x$ to the positive, boundary and negative region of each class label $y_i$, respectively. These membership degrees are computed from the inclusion degree of the similarity class of $x$ into each information granule,

\begin{equation}
\label{eq:rst:pos}
\mu_{\mathcal{P}_{(y_i)}}^\mathcal{R}(x) = \frac{|\bar{\mathcal{R}}(x) \cap \mathcal{P}(X_{[y_i]})|}{|\mathcal{P}(X_{[y_i]})|}
\end{equation}

\begin{equation}
\label{eq:rst:bnd}
\mu_{\mathcal{B}(y_i)}^\mathcal{R}(x) = \frac{|\bar{\mathcal{R}}(x) \cap \mathcal{B}(X_{[y_i]})|}{|\mathcal{B}(X_{[y_i]})|}
\end{equation}

\begin{equation}
\label{eq:rst:neg}
\mu_{\mathcal{N}(y_i)}^\mathcal{R}(x) = \frac{|\bar{\mathcal{R}}(x) \cap \mathcal{N}(X_{[y_i]})|}{|\mathcal{N}(X_{[y_i]})|}
\end{equation}

\noindent where $\bar{\mathcal{R}}(x)$ is the similarity class associated with the instance $x$, whereas $X_{[y_i]}$ denotes the set of instances with label $y_i$. The similarity class of an instance $x$ groups all instances that are similar to $x$ according to the subset of attributes taken into account. By computing how much $x$ and its similar instances are included in the positive region of a class $y_i$, we are estimating how sure we are of this classification. The same reasoning holds for the negative and boundary regions.

Equation \eqref{eq:rst:weight} computes the weight for the instance $x$ belonging to the enlarged dataset, given its label $y_i$ and a similarity relation $\mathcal{R}$. The sigmoid function $\varphi(.)$ is used to maintain the weight in the $(0,1)$ range.

\begin{equation}
\label{eq:rst:weight}
w_{(x,y_i)}= \varphi \left (\mu_{\mathcal{P}(y_i)}^\mathcal{R}(x) + 0.5 * \mu_{\mathcal{B}(y_i)}^\mathcal{R}(x) - \mu_{\mathcal{N}(y_i)}^\mathcal{R}(x) \right )
\end{equation}

The intuition of this weight is that if an instance and its similar ones are included mostly in the positive region of a class, therefore their label must be correct and they should have a high weight in the second learning phase. In the same way, if the instance $x$ and its similarity class are mostly contained in the negative region of a class then the class assigned to $x$ by the black box must be a mistake. Observe that the boundary information is also interesting since a high inclusion degree of an instance in the boundary region of a class is to some extent positive evidence (see Equation \ref{eq:rst:up}). This boundary region role can be reinforced or diluted according to the evidence coming from the inclusion degrees in the other two regions. When using the RST-based amending, Equation \eqref{eq:rst:weight} replaces Equation \eqref{eq:weightsunlab} in the pseudo-code of Algorithm \ref{alg}.

\begin{figure}[!htbp]
  \centering
  \includegraphics[height=5cm]{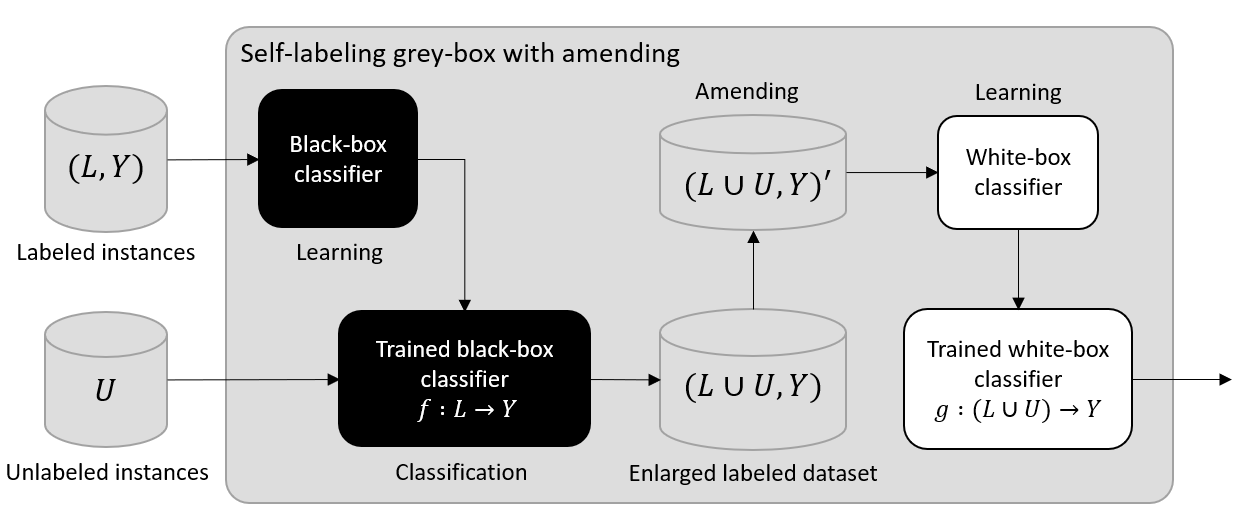} \\
  \caption{Blueprint of the SlGb architecture using amending procedures for correcting the influence of the misclassifications from the self-labeling process. When RST-based amending is used, it also tackles class inconsistency coming from noise in the labeled data.}
  \label{fig:greybox-mod2}
\end{figure}

Figure \ref{fig:greybox-mod2} illustrates the inclusion of the amending procedures into the learning algorithm of the SlGb approach. It is important to note that the amending process is only carried out in the learning phase of the SlGb. Therefore, the amending strategies do not affect the transparency of the white-box surrogate during the inference on new cases. 

The use of amending by weighting could have some implications for the interpretability. Assigning high weights to a small subset of instances transforms the global surrogate model towards a more local one. In other words, the weighting of instances makes the white box biased towards learning from the most confident ones, thus providing explanations for that subspace of the domain mostly. However, it makes sense to provide interpretability or explanations over the predictions that are most certain in the problem domain. In addition, it could have a positive influence on reducing the number of explanations produced by the white box.

\section{Experiments and Discussion}
\label{sec:exp}

In this section, we evaluate the SlGb approach through a three-step methodology using standard benchmark datasets. Unlike other experiments reported in the literature, the one developed in this section evaluates both algorithms' performance and interpretability, when having different percentages of labeled instances. As a complement, we propose three new evaluation measures that go beyond the prediction rates.

Being more specific, the \textit{first step} of our experimentation methodology is devoted to determining which black-box classifier produces the best results in terms of prediction performance. This step is quite important since the overall performance will depend on the discriminatory ability of the black box. The \textit{second step} is dedicated to determining which combination of white box and amending reaches the best commitment between prediction rates and interpretability. As a \textit{third step}, we further explore the impact of having different percentages of labeled and unlabeled instances on the algorithm's performance.

As a complement of the evaluation methodology for interpretable SSC methods, in the last part of this section we compare SlGb against the best-performing state-of-the-art methods. In this case, the evaluation is confined to the prediction rates as these methods cannot be interpreted, thus the goal here is to show that SlGb is not just simple and elegant, but also able to outperform other self-labeling methods reported in the literature.

\subsection{Benchmark Datasets, Base Classifiers and Parameter Settings}
\label{sec:exp:data}

Our experimental design includes 55 challenging and diverse datasets for classification tasks where features are structured (i.e. the dataset has tabular form) and therefore are potentially interpretable. Four ratios of labeled instances in the training set (from 10\% to 40\%) allow studying the influence of the number of labeled examples on the overall performance. Testing with a 10\% ratio means that the training set contains only a 10\% of labeled instances and the rest of are unlabeled, the instances in the test set are all labeled but set apart. These datasets comprise different characteristics: the number of instances ranges from 100 to 19000, the number of attributes from 2 to 90, and the number of decision classes from 2 to 28. Moreover, we have 25 datasets with different degrees of class imbalance and roughly half of the datasets are multiclass problems (see Table \ref{table:datasets}). 

These datasets are partitioned into training and test sets as done in a 10-fold cross-validation process, but each training set consists of labeled and unlabeled instances. The subset of unlabeled instances is obtained by performing a random selection without replacement and neglecting the class label of such instances. The ratio (10\% to 40\%) determines the number of labeled instances that are kept in this process for each training set. These datasets (including the cross-validation fold partitions) were provided as supplementary material in \cite{triguero2015self} and constitute an standard in the evaluation of shallow SSC techniques. We use these datasets, including the partitions as a form of guaranteeing a fair comparison against state-of-the-art SSC methods.

There are several algorithms that can be adopted as base classifiers. On one hand, the selected classifier for the base black box should exhibit a strong predictive capability as it is used to determine the decision class of unlabeled instances. Next, we describe three mainstream supervised classifiers that will be used in the experiments for instantiating the black-box component. Our choice is motivated by experimental evidence of their superior performance in a wide range of classification problems \cite{zhang2017up,fernandez2014we,wainberg2016random} and their ability to produce calibrated probabilities (except for support vector machines where a calibration post-hoc is needed).

\vspace{2mm}

\textbf{Black-box classifiers}

\begin{itemize}
	\item{Random Forests (RF) \cite{Breiman2001}: Ensemble of decision trees that uses bagging technique for aggregating the results in order to reduce the high variance of individual decision trees. Individual decision trees are built with a random subset of attributes and a random sample with replacement of instances. In our implementation 100 trees are aggregated and the number of random attributes to consider for each tree equals $\log_2(|A|)$.}
	\item{Multilayer Perceptron (MLP) \cite{Hecht-Nielsen1989}: Feed-forward neural network using backpropagation algorithm for adjusting its weights. Our implementation uses learning rate equals to 0.3, momentum equals to 0.2, 500 epochs for learning and one hidden layer with $(|A| + |Y|)/2$ as the number of neurons.}
	\item{Support Vector Machine (SVM) \cite{Platt1998,Keerthi2001}: Support vector machine classifier using sequential minimal optimization algorithm for training. Our implementation uses a polynomial kernel with Platt's scaling (logistic) calibration of probabilities.}
\end{itemize}

On the other hand, for the white-box component any intrinsically interpretable classifier can be used as a surrogate model. Therefore, the choice of a white box must be driven by the type of explanations that are desired, e.g. rules, feature coefficients, probabilities, examples, etc. We decide to explore decision trees and decision lists alternatives as they provide both intuitive individual explanations in the form of \textit{if-then} rules and a view of the model as a whole. For decision trees, the hierarchical structure provides this view and it can be considered transparent as long as the size of the tree remains manageable. For the case of the decision lists, rules sets are generally more concise than the ones extracted from decision trees. Additionally, these algorithms are able to handle weighted instances in the learning process. Next, we describe three classifiers explored in the scope of this experiment.

\vspace{2mm}

\textbf{White-box classifiers}

\begin{itemize}
	\item{Decision Tree (C45) \cite{quinlan1993}: Our implementation uses C4.5 algorithm for inducing a decision tree. We allow two instances as the minimum number of instances per leaf. The confidence factor for pruning is 0.25, where a lower value incurs in more pruning. When pruning the sub-tree raising operation is used.}
	\item{PART Decision List (PART) \cite{Frank1998}: PART uses the separate-and-conquer strategy for building a rule set by generating a partial C4.5 decision tree and making the most confident leaf into a rule. In the next iteration, all covered instances are removed from the dataset and the process is repeated. Thus, decision lists must be interpreted in order. Our implementation uses the same hyper-parameters of the decision tree described above for generating the partial C4.5 decision trees.}
	\item{RIPPER Decision List (RIP) \cite{cohen1995}: This method is a propositional rule learner with a separate-and-conquer strategy, as described for PART. Additionally, the training data is split into a growing set and a pruning set for performing reduced error pruning. The rule set formed from the growing set is simplified with pruning operations optimizing the error on the pruning set. For our implementation, the minimum allowed support of a rule is two and the data is split in three folds where one is used for pruning. Besides, two optimization iterations are performed.}
\end{itemize}

For completeness, we enumerate the amending procedures that will be tested in combination with the previous base classifiers. 

\subsubsection{Amending procedures}

\begin{itemize}
	\item{No amending (NONE): The first option is not using amending. All self-labeled instances are provided as extra data to the surrogate white box. This is used as a baseline for evaluating the contribution of the two amending procedures proposed.}
	\item{Amending based on class membership probabilities (CONF): Amending procedure based on calibrated class membership probabilities obtained from the black-box base classifier \cite{grau2018}}.
	\item{Amending based on RST inclusion degree measure (RST): Amending procedure based on RST aiming to correct the inconsistency in the classifications, as described in the previous section}.
\end{itemize}
	
Hereinafter, when referring to a particular configuration of SlGb we denote it as ``\textit{bb-wb-am}" where \textit{bb} represents the base black box, \textit{wb} represents the surrogate white box and \textit{am} represents the amending procedure\footnote{Code, datasets and results using different measures (kappa, accuracy, number of rules, etc.) are available as supplementary material for reproducibility purposes in \textcolor{blue}{\url{gitlab.ai.vub.ac.be/igraugar/slgb_scripts/tree/paper}}. All SlGb configurations were implemented using \textit{weka} library \cite{Witten2017467} and its default parameters listed above.}.

\subsection{Impact of the Black-box Base Classifiers on the Performance}
\label{sec:exp:black}

We first focus on evaluating the influence of the base black box in the performance of the algorithm. Here no amending procedure is taken into account yet since it does not directly affects the ability of the black box to produce correct classifications. In order to measure the configurations in terms of prediction rates we report the Cohen's kappa coefficient \cite{kappa1960}. This measure estimate the inter-rater agreement for categorical items and ranges in $[-1,1]$, where $-1$ indicates no agreement between the prediction and the actual values, $0$ means no learning (i.e., random prediction), and $1$ total agreement or perfect performance. While accuracy is considered mainstream when measuring classification rates, the kappa is a more robust measure since this coefficient takes into account the agreement occurring by chance, which is especially relevant for datasets with class imbalance \cite{japkowicz2011evaluating,ben2008comparison}.

Table \ref{table:performance:noamend} gives the mean and the standard deviation of the kappa coefficient achieved on each setting. We group the results for different percentages of labeled instances. The numerical simulations indicate that using RF as the black-box component leads to higher prediction rates. In particular, the RF-PART-NONE configuration stands as the best performing one for varying amounts of labeled instances, very closely followed by RF-C45-NONE and RF-RIP-NONE.

\begin{table}[!ht]
\centering
\caption{Prediction rates (kappa) achieved by different combinations of black-box and white-box algorithms without using amending. Results are grouped by ratio and best results are highlighted in bold.}
\label{table:performance:noamend}
\begin{tabular}{|c|c|cccc|}
\thickhline
                          & Ratio & 10\%   & 20\%   & 30\%   & 40\%   \\ \thickhline
\multirow{2}{*}{MLP-C45-NONE}   & mean  & 0.50 & 0.53 & 0.55 & 0.56 \\ \cline{2-6}
                                & stdev & 0.28 & 0.28 & 0.28 & 0.28 \\ \hline
\multirow{2}{*}{MLP-PART-NONE}  & mean  & 0.50 & 0.54 & 0.56 & 0.57 \\ \cline{2-6} 
                                & stdev & 0.29 & 0.28 & 0.28 & 0.28 \\ \hline
\multirow{2}{*}{MLP-RIP-NONE}   & mean  & 0.51 & 0.54 & 0.55 & 0.57 \\ \cline{2-6} 
                                & stdev & 0.29 & 0.28 & 0.28 & 0.28 \\ \thickhline
\multirow{2}{*}{RF-C45-NONE}    & mean  & \textbf{0.56} & \textbf{0.60} & \textbf{0.61} & 0.61 \\ \cline{2-6} 
                                & stdev & 0.28 & 0.27 & 0.27 & 0.27 \\ \hline
\multirow{2}{*}{RF-PART-NONE}   & mean  & 0.56 & \textbf{0.60} & \textbf{0.61} & \textbf{0.62} \\ \cline{2-6} 
                                & stdev & 0.29 & 0.27 & 0.27 & 0.27 \\ \hline
\multirow{2}{*}{RF-RIP-NONE}    & mean  & 0.55 & \textbf{0.60} & \textbf{0.61} & \textbf{0.62} \\ \cline{2-6} 
                                & stdev & 0.28 & 0.27 & 0.27 & 0.27 \\ \thickhline
\multirow{2}{*}{SVM-C45-NONE}   & mean  & 0.49 & 0.53 & 0.55 & 0.56 \\ \cline{2-6} 
                                & stdev & 0.28 & 0.27 & 0.27 & 0.26 \\ \hline
\multirow{2}{*}{SVM-PART-NONE}  & mean  & 0.50 & 0.53 & 0.56 & 0.57 \\ \cline{2-6} 
                                & stdev & 0.28 & 0.28 & 0.28 & 0.27 \\ \hline
\multirow{2}{*}{SVM-RIP-NONE}   & mean  & 0.50 & 0.53 & 0.55 & 0.57 \\ \cline{2-6} 
                                & stdev & 0.28 & 0.28 & 0.27 & 0.27 \\ \thickhline
\end{tabular}
\end{table}
         
With the aim of providing a rigorous statistical analysis of the differences, we compute the Friedman two-way analysis of variances by ranks \cite{Friedman1937}, per ratio. The test suggests rejecting the null hypotheses for all labeled ratios based on a confidence interval of 95\% (see Table \ref{table:friedman1} in appendix\footnote{All tables related to statistical tests are included in the appendix.}). This means that there exist significant differences between at least two configurations on each ratio. 

The next step is focused on determining whether RF black box is truly superior compared to other configurations. To do so, we adopt the Wilcoxon signed rank test \cite{Wilcoxon1945} and Holm's post-hoc procedure \cite{holm1979simple} to correct the $p$-values, as suggested by Benavoli \textit{et al.} \cite{Benavoli2016}. Table \ref{table:wilcoxon1} reports the unadjusted $p$-value computed by the Wilcoxon test and the corrected $p$-value associated with each pairwise comparison. In order to discover the influence of the black box we compare the pairs of configurations using the same surrogate white box. Each section of the table represents the ratio of labeled instances. The null hypothesis states that there is no significant difference between the performance of each pair of configurations. All null hypotheses are rejected, except for RF-RIP vs. MLP-RIP in the 40\% ratio (however RF still has higher prediction rates). 

This suggests that RF is clearly the best-performing base black box for the self-learning grey-box. This result is not surprising since RF has proven to be very competent classifier in different experimental studies \cite{zhang2017up,fernandez2014we,wainberg2016random,touw2012data}. Furthermore, RF produces consistent probabilities that does not need to be calibrated \cite{niculescu2005predicting}, which is a requirement for the later use of confidence-based amending.

\subsection{Impact of using Different White Boxes and Amending Configurations}
\label{sec:exp:white}

In this section, we study how different choices of the amending processes and white-box surrogates impact the overall results. We propose some measures for evaluating performance taking both accuracy and interpretability into account.

We first explore the influence on the prediction rates. Based on the selection of RF as black-box base, Table \ref{table:performance:amend} shows very similar results across each ratio. Going deeper with the statistical analysis, we apply Friedman and Wilcoxon tests with post-hoc correction. Although Friedman test finds significant differences in the four groups (Table \ref{table:friedman2}), examining Wilcoxon corrected tests we ascertain that the null hypothesis cannot be rejected for the vast majority of pairs compared (see Tables \ref{table:wilcoxon2a} and \ref{table:wilcoxon2b} for details). This means that there are no statistically significant differences in the prediction rates when comparing different amending procedures with a fixed white box and vice versa. This behavior suggests that the overall prediction rates of the approach mostly relies on the correct choice of the black-box algorithm.

\begin{table}[!ht]
\centering
\caption{Prediction rates achieved by different combinations of white boxes and amending strategies while using RF as black box. Results are grouped by ratio and best results are highlighted in bold.}
\label{table:performance:amend}
\begin{tabular}{|c|c|c|c|c|c|}
\thickhline
                              & Ratio & 10\%   & 20\%   & 30\%   & 40\%   \\ \thickhline
\multirow{2}{*}{RF-C45-NONE}  & mean  & \textbf{0.56} & 0.60 & 0.61 & 0.61 \\ \cline{2-6} 
                              & stdev & 0.28 & 0.27 & 0.27 & 0.27 \\ \hline
\multirow{2}{*}{RF-PART-NONE} & mean  & \textbf{0.56} & 0.60 & 0.61 & \textbf{0.62} \\ \cline{2-6} 
                              & stdev & 0.28 & 0.27 & 0.27 & 0.27 \\ \hline
\multirow{2}{*}{RF-RIP-NONE}  & mean  & 0.55 & 0.60 & 0.61 & \textbf{0.62} \\ \cline{2-6} 
                              & stdev & 0.28 & 0.27 & 0.27 & 0.27 \\ \thickhline
\multirow{2}{*}{RF-C45-CONF}  & mean  & 0.55 & 0.59 & 0.61 & 0.61 \\ \cline{2-6} 
                              & stdev & 0.29 & 0.28 & 0.28 & 0.28 \\ \hline
\multirow{2}{*}{RF-PART-CONF} & mean  & 0.56 & 0.60 & 0.61 & \textbf{0.62} \\ \cline{2-6} 
                              & stdev & 0.29 & 0.27 & 0.27 & 0.27 \\ \hline
\multirow{2}{*}{RF-RIP-CONF}  & mean  & 0.54 & 0.59 & 0.61 & 0.60 \\ \cline{2-6} 
                              & stdev & 0.29 & 0.27 & 0.27 & 0.28 \\ \thickhline
\multirow{2}{*}{RF-C45-RST}   & mean  & 0.56 & 0.60 & \textbf{0.62} & 0.62 \\ \cline{2-6} 
                              & stdev & 0.29 & 0.27 & 0.27 & 0.28 \\ \hline
\multirow{2}{*}{RF-PART-RST}  & mean  & \textbf{0.56} & \textbf{0.61} & \textbf{0.62} & \textbf{0.62} \\ \cline{2-6} 
                              & stdev & 0.28 & 0.27 & 0.27 & 0.27 \\ \hline
\multirow{2}{*}{RF-RIP-RST}   & mean  & 0.53 & 0.57 & 0.58 & 0.59 \\ \cline{2-6} 
                              & stdev & 0.28 & 0.28 & 0.28 & 0.28 \\ \thickhline
\end{tabular}
\end{table}

\begin{figure}[!htbp]
\begin{subfigure}{\textwidth}
\centering
\includegraphics[height=4.0cm]{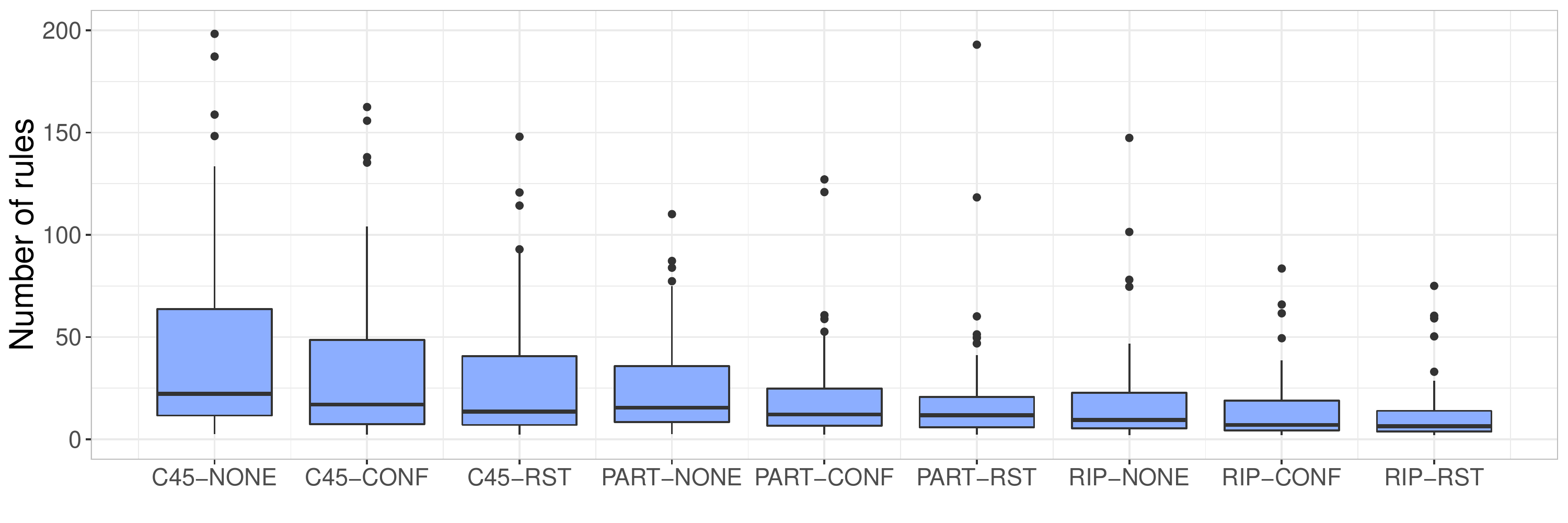}
\caption{Using 10\% of labeled instances.}
\end{subfigure}
\begin{subfigure}{\textwidth}
\centering
\includegraphics[height=4.0cm]{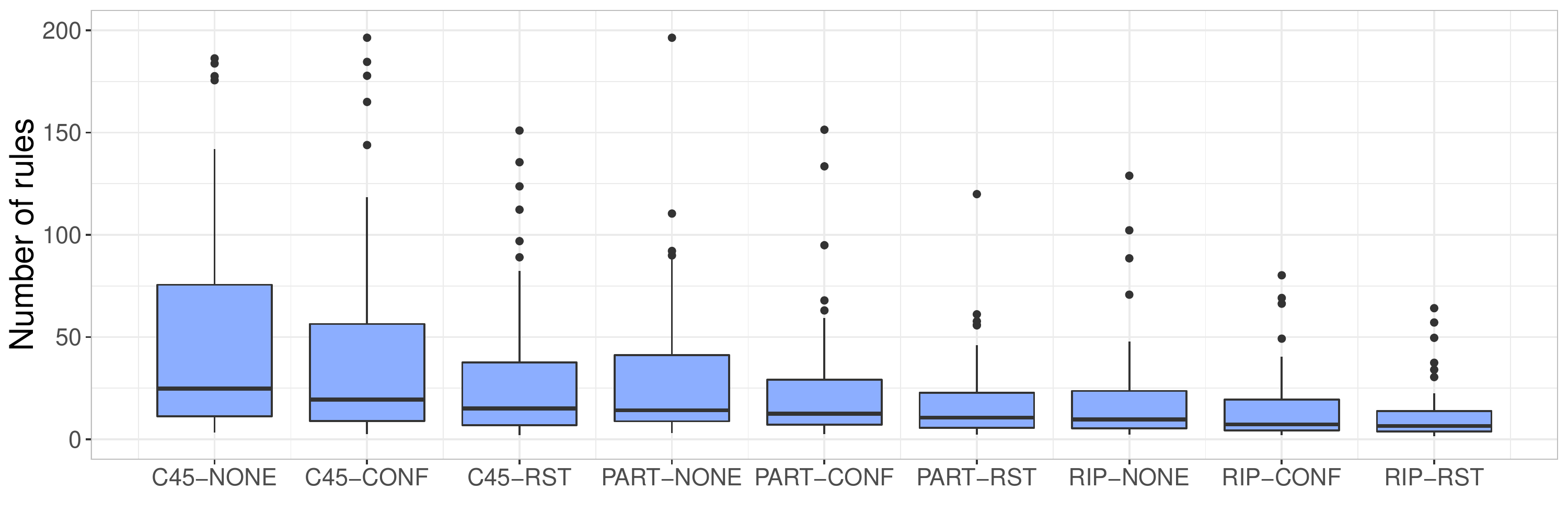}
\caption{Using 20\% of labeled instances.}
\end{subfigure}
\begin{subfigure}{\textwidth}
\centering
\includegraphics[height=4.0cm]{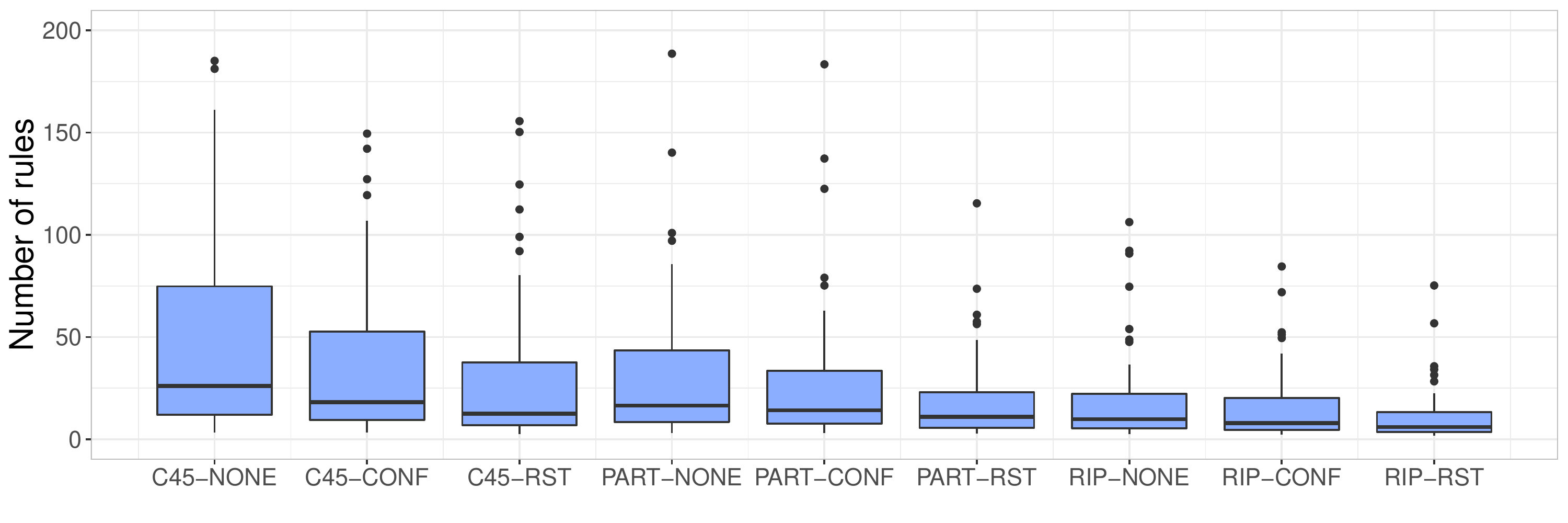}
\caption{Using 30\% of labeled instances.}
\end{subfigure}
\begin{subfigure}{\textwidth}
\centering
\includegraphics[height=4.0cm]{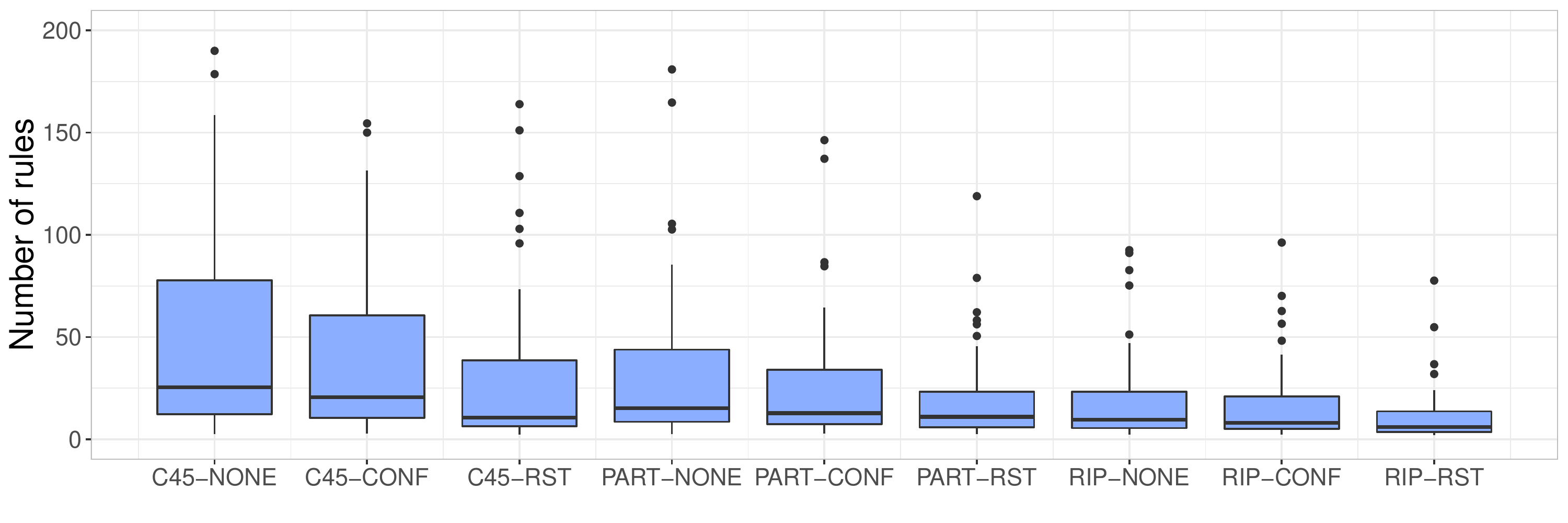}
\caption{Using 40\% of labeled instances.}
\end{subfigure}
\caption{Number of rules produced by each combination of white box and amending, using random forests as black box. Both amending strategies (specially RST) reduce the number of rules while RIP white box produces the lowest number of rules.}
\label{fig:box_amending}
\end{figure}

However, when examining the number of rules obtained, the difference is significantly visible. Figure \ref{fig:box_amending} plots of the number of rules produced by each combination, per ratio of labeled data. Two results are consistent across ratios: both amending strategies (specially RST) reduce the number of rules while RIP as surrogate white box produces the lowest number of rules for all possible combinations.

Toward exploring this result further, we also propose two new measures to evaluate models' interpretability via a quantifiable proxy. The first measure can be used in the context of self-labeling and the second one is applicable to any model containing explanation units. According to \cite{doshi2017towards}, there are three main forms of evaluating interpretability: application-grounded, human-grounded and functionally-grounded metrics. The functionally-grounded approach is the only one not requiring human experiments and collaboration. As an alternative it uses desiderata for interpretability (e.g. transparency, trust, etc.) as a proxy for assessing the quality of the model. Since we are working with benchmark datasets, we use the functionally-grounded approach for creating measures based on the simplicity as a mean for gaining transparency and simulatability (i.e. a human is able to simulate and reason about the model's entire decision-making process). The first measure can be used in the context of self-labeling for base methods that produce tree structures, rules or decision lists. It involves the number of rules in the decision lists (or equivalently the number of leaves in a decision tree) and expresses the \textit{relative growth} in structure as:

\begin{equation}
\label{eq:exp:growth}
\Gamma = |E^g| / |E^w| 
\end{equation}

\noindent where $E^g$ is the set of rules produced by the self-labeling method (here the grey-box) and $E^w$ is the set of rules produced by the baseline white box when using only labeled data. For this measure, a number much greater than one indicates that a major growth in the structure of the self-labeling method is needed when using the extra unlabeled data. In that case, the balance between interpretability and performance must be taken into account for further evaluation.

\begin{figure}[!htbp]
  \centering
  \includegraphics[height=5cm]{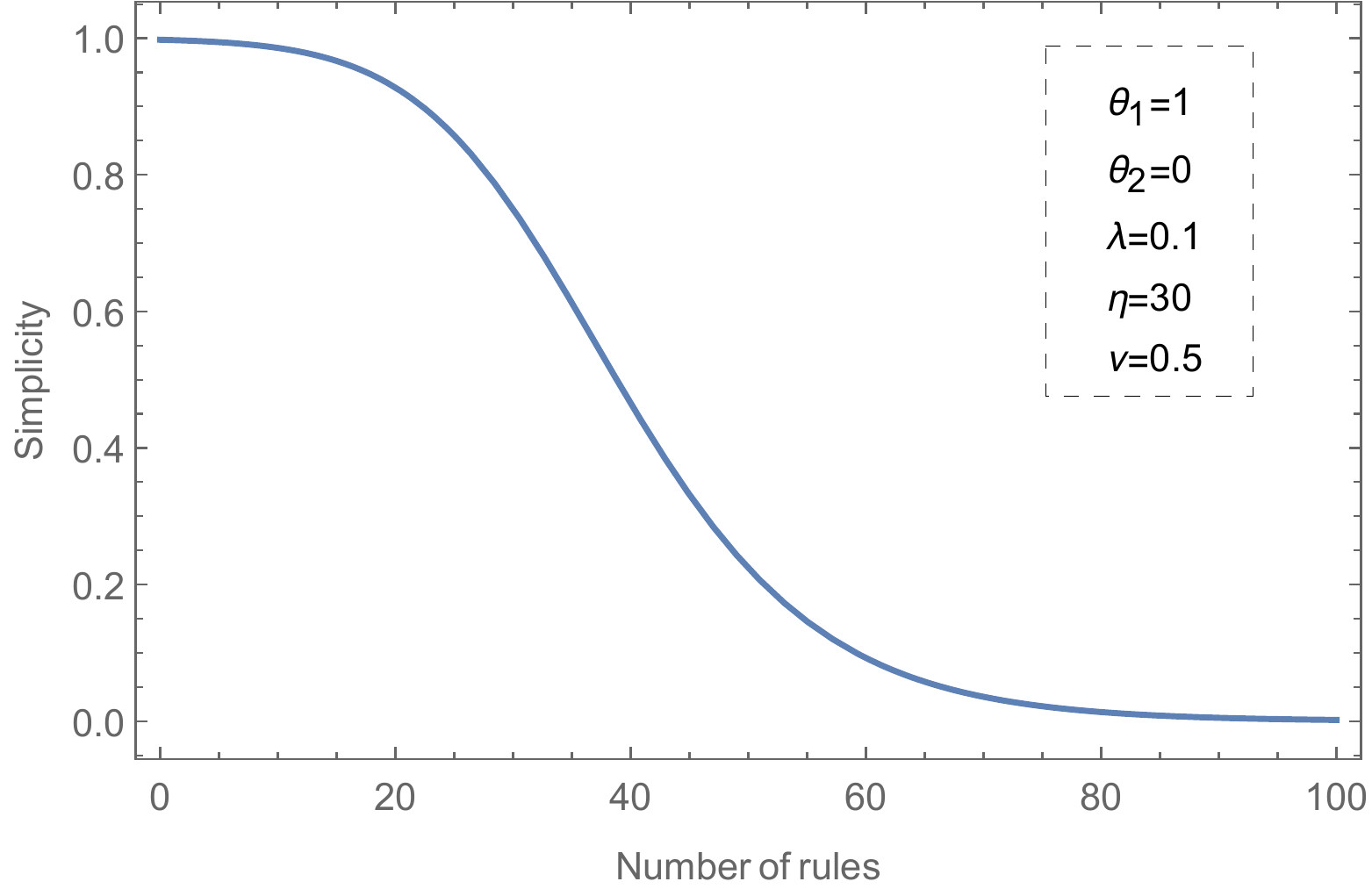} \\
  \caption{Simplicity function with default parameters used for the benchmark datasets. For specific applications these parameters are domain dependent.}
  \label{fig:simp}
\end{figure}

The second measure is more general and applicable to any model whose structure is formed by quantifiable explanation units (e.g. rules, prototypes, features, derived features, etc.). For our case, this measure estimates the \textit{simplicity} of the model according to the size of the structure in terms of the number of rules. Although the first notion would be that the smaller the rule set the better, this is not necessarily a linear relation. The desired simplicity in terms of the number of rules has a smooth behavior which can drop quickly. Therefore, we propose to measure simplicity through a generalized sigmoid function which has been historically used for fitting growth curves \cite{birch1999new}, since it allows representing this relation with enough flexibility. The simplicity can be formalized as the following equation:

\begin{equation}
\label{eq:exp:simplicity}
\Upsilon(|E^g|) = \theta_1 + \frac{\theta_2-\theta_1}{(1+e^{-\lambda(|E^g|-\eta)})^{1/\nu}}
\end{equation}

\noindent where $\theta_1 = 1$ and $\theta_2 = 0$ represent the upper and lower asymptotes of the function respectively, $\lambda$ is the slope of the curve, $\eta$ regulates the shift over the $x$-axis and $\nu$ affects near which asymptote maximum growth occurs. In this way, a result value of one indicates high simplicity and it decreases smoothly towards zero. A bigger $\lambda$ would make the function less smooth, generating a drastic drop in simplicity after a threshold in the number of rules is surpassed. The value of $\eta$ defines where the middle value of the function is obtained. While a value of $\nu=1$ makes no change in the curve, $\nu<1$ moves the growth toward the upper asymptote and $\nu>1$ toward the lower one. Observe that both $\eta$ and $\nu$ influence where 0.5 simplicity is obtained. Given the diversity of our benchmark data, we take $\lambda=0.1, \eta=30, \nu=0.5$ for illustrating a general setting (see Figure \ref{fig:simp}). 

With these values, the function produces medium evaluations (around 0.5) when the number of rules is around 40. Similarly, it obtains rather high simplicity (higher than 0.8) when the number of rules goes below 30. However, parameter values should be estimated based on expert knowledge for specific applications. This highly flexible function allows customizing the value of simplicity according with the specifics of a given case study.

Table \ref{table:performance:interp} shows the average relative growth and  simplicity over the 55 datasets tested for the four ratios. Regarding the relative growth, the increase in the structure of the grey-box is on average larger when using small amounts of labeled data, while for bigger ratios this difference decreases. This growth in the structure is an expected consequence of providing more unlabeled data to the white-box surrogate in the grey-box scheme. However, the use of amending procedures alleviates this effect by giving more importance to relevant unlabeled instances. In general, a smaller growth is observed when using RST amending, especially in combination with PART as the white box, thus resulting in the winner combination for all ratios.

\begin{table}[!ht]
\centering
\caption{Mean (and standard deviation) of the relative growth and simplicity achieved by different combinations of white boxes and amending strategies while using RF as black box. Results are grouped by ratio and best results are highlighted in bold.}
\label{table:performance:interp}
\begin{tabular}{|c|c|c|c|c|c|}
\thickhline
                              & Ratio      & 10\% & 20\% & 30\% & 40\% \\ \thickhline
\multirow{2}{*}{RF-C45-NONE}  & growth     & 4.24 (2.98) & 3.22 (3.81) & 3.00 (6.61) & 3.95 (15.12) \\ \cline{2-6} 
                              & simplicity & 0.57 (0.44) & 0.56 (0.45) & 0.56 (0.45) & 0.56 (0.45) \\ \hline
\multirow{2}{*}{RF-PART-NONE} & growth     & 3.07 (0.92) & 2.19 (0.62) & 1.78 (0.51) & 1.55 (0.47) \\ \cline{2-6} 
                              & simplicity & 0.70 (0.39) & 0.70 (0.39) & 0.69 (0.40) & 0.69 (0.40) \\ \hline
\multirow{2}{*}{RF-RIP-NONE}  & growth     & 3.93 (4.78) & 3.19 (4.13) & 2.93 (4.48) & 2.67 (4.37) \\ \cline{2-6} 
                              & simplicity & 0.85 (0.28) & 0.84 (0.29) & 0.84 (0.30) & 0.84 (0.30) \\ \thickhline
\multirow{2}{*}{RF-C45-CONF}  & growth     & 2.74 (2.38) & 2.34 (3.35) & 2.43 (5.96) & 3.39 (13.75) \\ \cline{2-6} 
                              & simplicity & 0.67 (0.42) & 0.63 (0.44) & 0.61 (0.45) & 0.60 (0.45) \\ \hline
\multirow{2}{*}{RF-PART-CONF} & growth     & 2.11 (0.59) & 1.66 (0.47) & 1.45 (0.43) & 1.30 (0.40) \\ \cline{2-6} 
                              & simplicity & 0.81 (0.32) & 0.78 (0.34) & 0.75 (0.35) & 0.74 (0.36) \\ \hline
\multirow{2}{*}{RF-RIP-CONF}  & growth     & 2.96 (2.94) & 2.52 (3.17) & 2.41 (3.94) & 2.54 (5.87) \\ \cline{2-6} 
                              & simplicity & 0.89 (0.39) & 0.88 (0.25) & 0.87 (0.26) & 0.86 (0.27) \\ \thickhline
\multirow{2}{*}{RF-C45-RST}   & growth     & 2.26 (0.93) & 1.53 (0.61) & 1.20 (0.59) & 1.00 (0.33) \\ \cline{2-6} 
                              & simplicity & 0.71 (0.39) & 0.71 (0.39) & 0.71 (0.39) & 0.71 (0.40) \\ \hline
\multirow{2}{*}{RF-PART-RST}  & growth     & \textbf{1.99 (0.49)} & \textbf{1.38 (0.31)} & \textbf{1.13 (0.24)} & \textbf{0.98 (0.21)} \\ \cline{2-6} 
                              & simplicity & 0.82 (0.32) & 0.81 (0.33) & 0.81 (0.33) & 0.81 (0.34) \\ \hline
\multirow{2}{*}{RF-RIP-RST}   & growth     & 2.42 (2.26) & 1.69 (1.06) & 1.39 (0.63) & 1.20 (0.42) \\ \cline{2-6} 
                              & simplicity & \textbf{0.91 (0.23)} & \textbf{0.92 (0.21)} & \textbf{0.93 (0.19)} & \textbf{0.94 (0.18)} \\ \thickhline
\end{tabular}
\end{table}

In addition, the simplicity measure (the closer the value to one the better) also indicates that the use of amending is convenient for obtaining more concise sets of rules. It is also evident that using RIP as surrogate generates the least number of rules, followed by PART. For this measure the absolute winner is RF-RIP-RST combination, exhibiting the highest values of simplicity for all ratio values used for experimentation. A similar statistical validation support this statement (see tables \ref{table:friedman3} and \ref{table:wilcoxon3}), finding significant statistical differences when comparing RF-RIP-RST with other configurations using simplicity as interpretability measure.

It is important to remark that the simplicity measure solely quantifies what it would be considered a simulatable model. Of course, a very simple model with only one rule and poor prediction rates is not desirable, whereas for a very simple dataset three or four rules might be enough to reach accurate results. That is why taking into account the prediction performance is fundamental for a proper assessment. To measure algorithms' quality based on the balance between the prediction rates and the simplicity of the learned model, we propose a third measure, called \textit{utility}, combining the kappa (re-scaled to (0,1)) and the simplicity values with a weighing parameter $\alpha$, 

\begin{equation}
\label{eq:exp:usefulness}
\Psi(E^g) = \alpha * \kappa(E^g)' + (1-\alpha)* \Upsilon(|E^g|) 
\end{equation}

\noindent where $\alpha$ is set to 0.6 in our experimental setting, representing a scenario where the accuracy and the interpretability have almost the same preference. Utility functions are commonly used in multi-objective optimization for mapping a vector of pay-offs to a single scalar value \cite{Radulescu2020}. In this case, the utility function is a linear combination of two terms parameterized by the weight $\alpha$. This weighing parameter allows adjusting the preference of the user for prioritizing the accuracy or the interpretability objectives. Here, the two objectives are measured based on kappa and simplicity, respectively. It would be interesting to extend the proposed utility to involve more objectives where the parameters should be obtained from the preferences of a panel of domain experts \cite{Zintgraf2018,Roijers2013}.

\begin{figure}[!htbp]
\begin{subfigure}{0.5\textwidth}
\centering
\includegraphics[height=5cm]{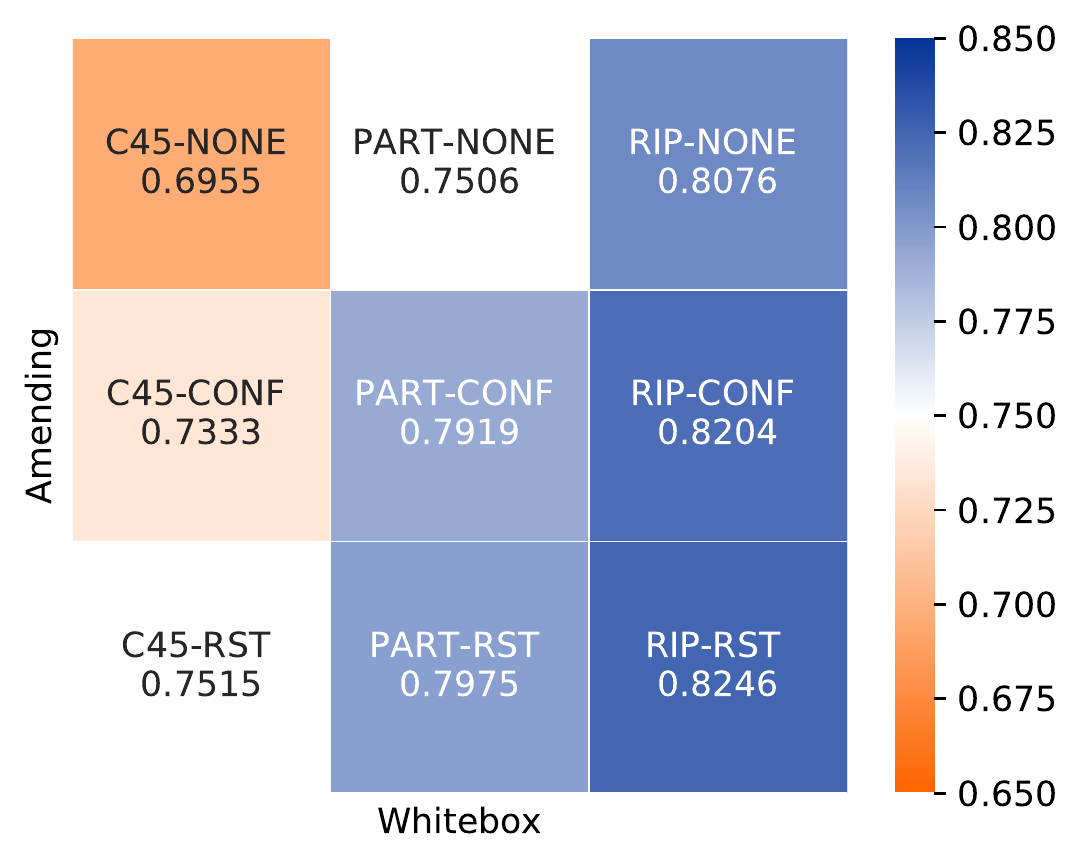}
\caption{Using 10\% of labeled instances.}
\end{subfigure}
\begin{subfigure}{0.5\textwidth}
\centering
\includegraphics[height=5cm]{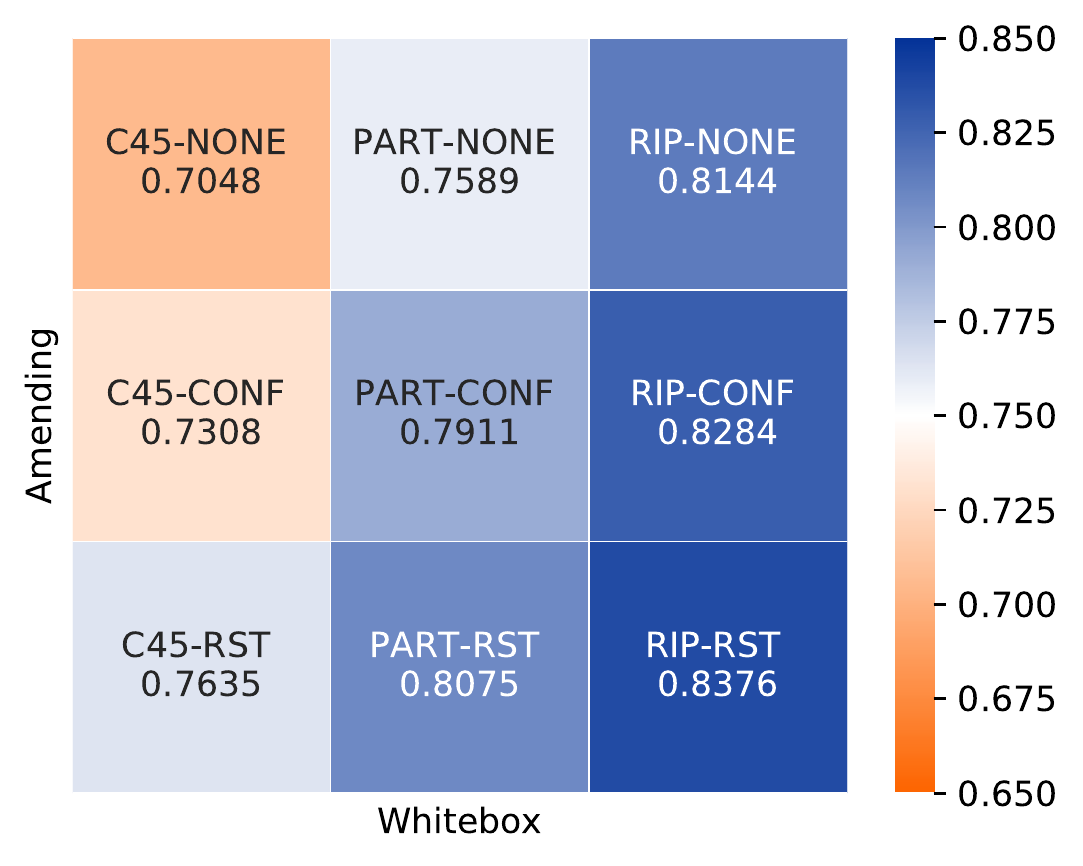}
\caption{Using 20\% of labeled instances.}
\end{subfigure}
\begin{subfigure}{0.5\textwidth}
\centering
\includegraphics[height=5cm]{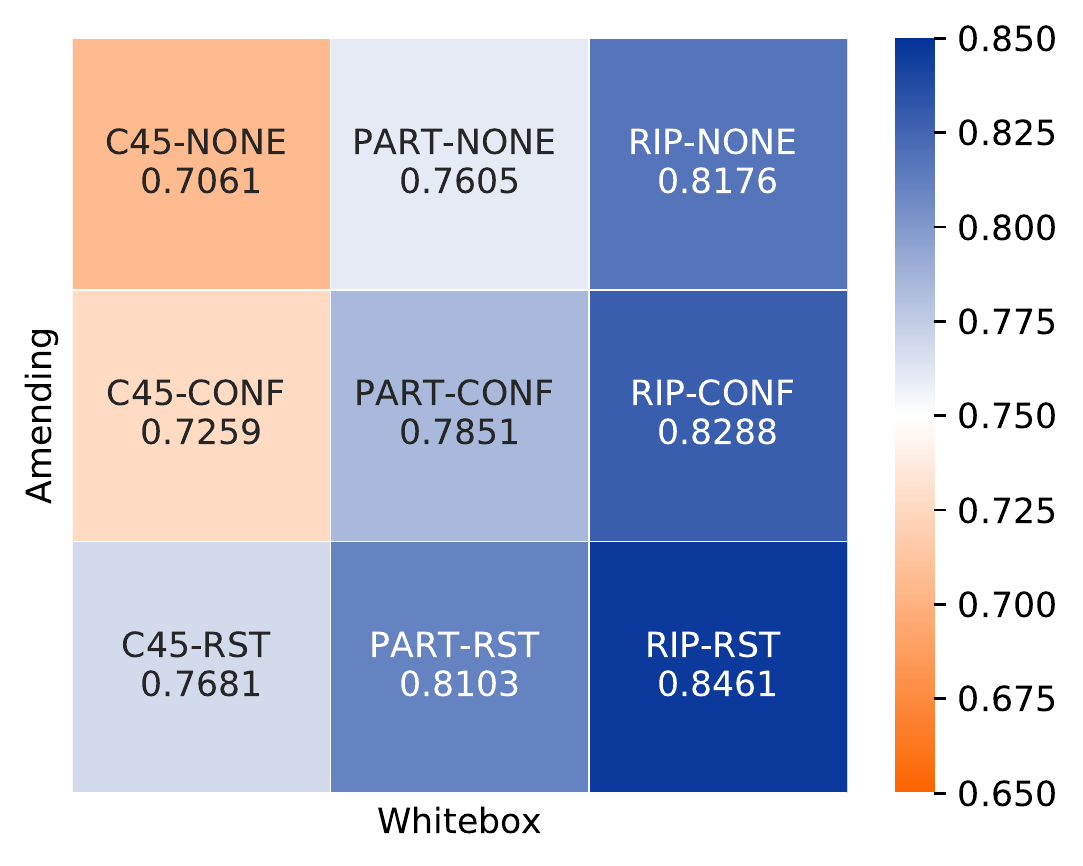}
\caption{Using 30\% of labeled instances.}
\end{subfigure}
\begin{subfigure}{0.5\textwidth}
\centering
\includegraphics[height=5cm]{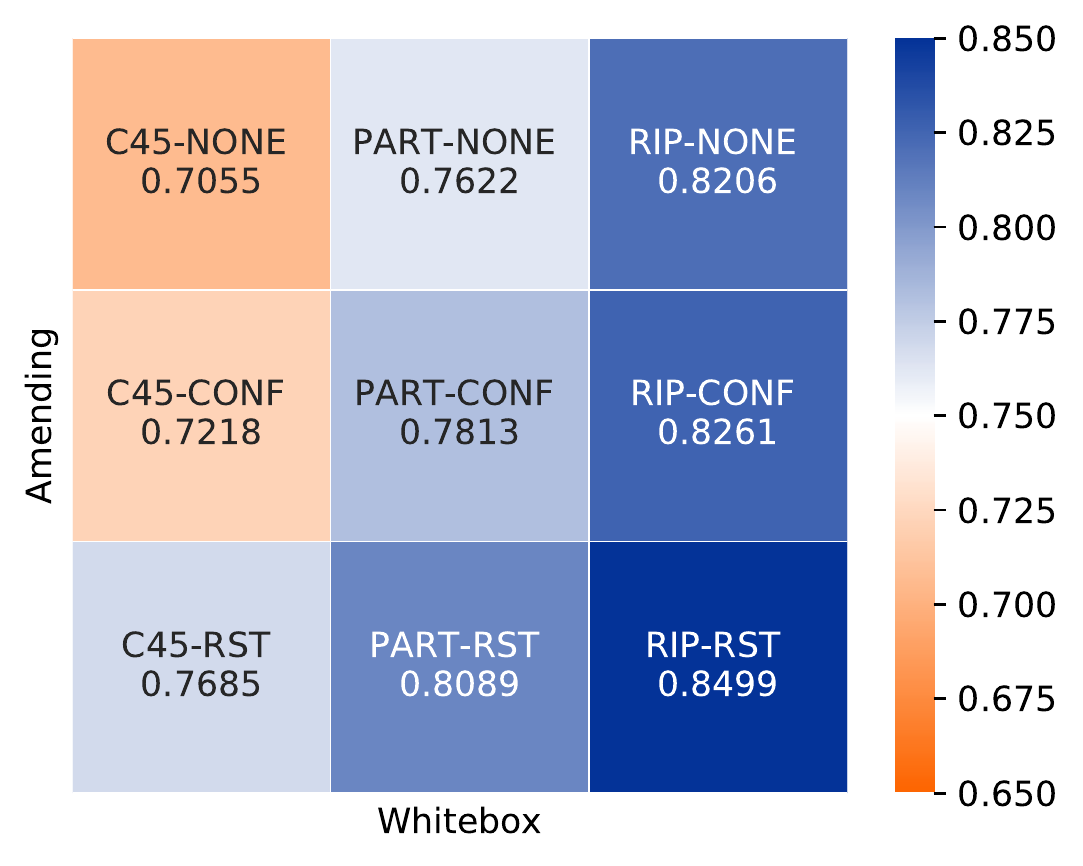}
\caption{Using 40\% of labeled instances.}
\end{subfigure}
\caption{Mean utility values of each combination of white box and amending, using random forests as the black-box base classifier. The use of RIP as a white-box component in combination with the RST-based amending achieves the best trade-off between accuracy and interpretability, for all explored ratios.}
\label{fig:utility}
\end{figure}

As a partial summary, Figure \ref{fig:utility} visualizes the utility values in a heat-map plot. From this figure, it is easy to perceive that the RIP algorithm, as a white-box surrogate, positively contributes to the overall performance of the approach when taking both kappa and simplicity into account. Additionally, RST amending also increases the value of utility when compared with CONF amending or not using amending at all. This measure reflects that, in general, the best trade-off is reached when using the RF-RIP-RST combination and the highest values are achieved when more labeled data is available.

\subsection{Influence of the Number of Labeled and Unlabeled Instances}
\label{sec:exp:size}

In this subsection, we use RF-RIP-RST to explore the impact of having different amounts of labeled and unlabeled instances on algorithm's results. In the evaluation of semi-supervised techniques, it is a common strategy to vary the size of $L$ by systematically neglecting the label of different amounts of instances and adding them to $U$. But this procedure does not explore the scenario where also the unlabeled instances could be hard to obtain \cite{avital2018realistic}. Observe that since this is a controlled experiment we can safely assume that the unlabeled instances follow the same class distribution as the labeled ones. In reality, one might need to re-balance the dataset after self-labeling if the unlabeled instances per-class distribution significantly differs.

\begin{figure}[!htbp]
\begin{subfigure}{0.5\textwidth}
\centering
\includegraphics[height=4.75cm]{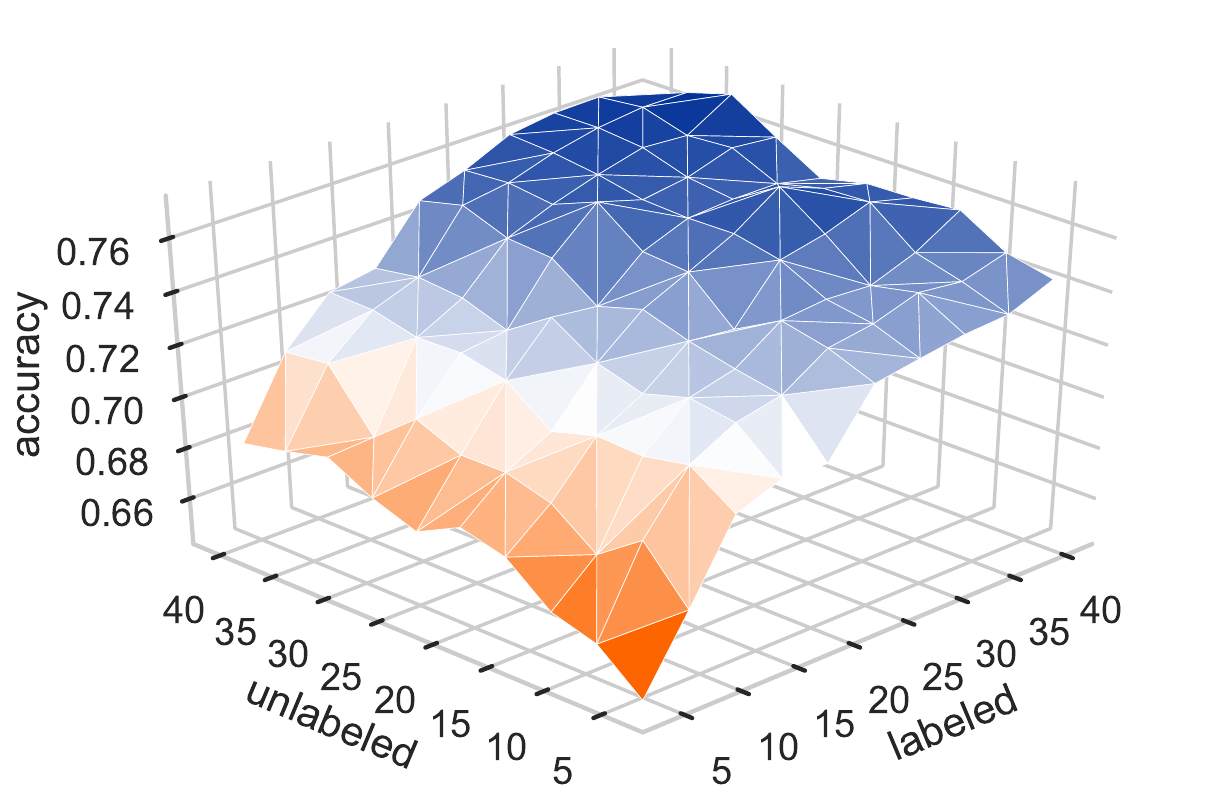}
\caption{Accuracy.}
\label{fig:surfacea}
\end{subfigure}
\begin{subfigure}{0.5\textwidth}
\centering
\includegraphics[height=4.75cm]{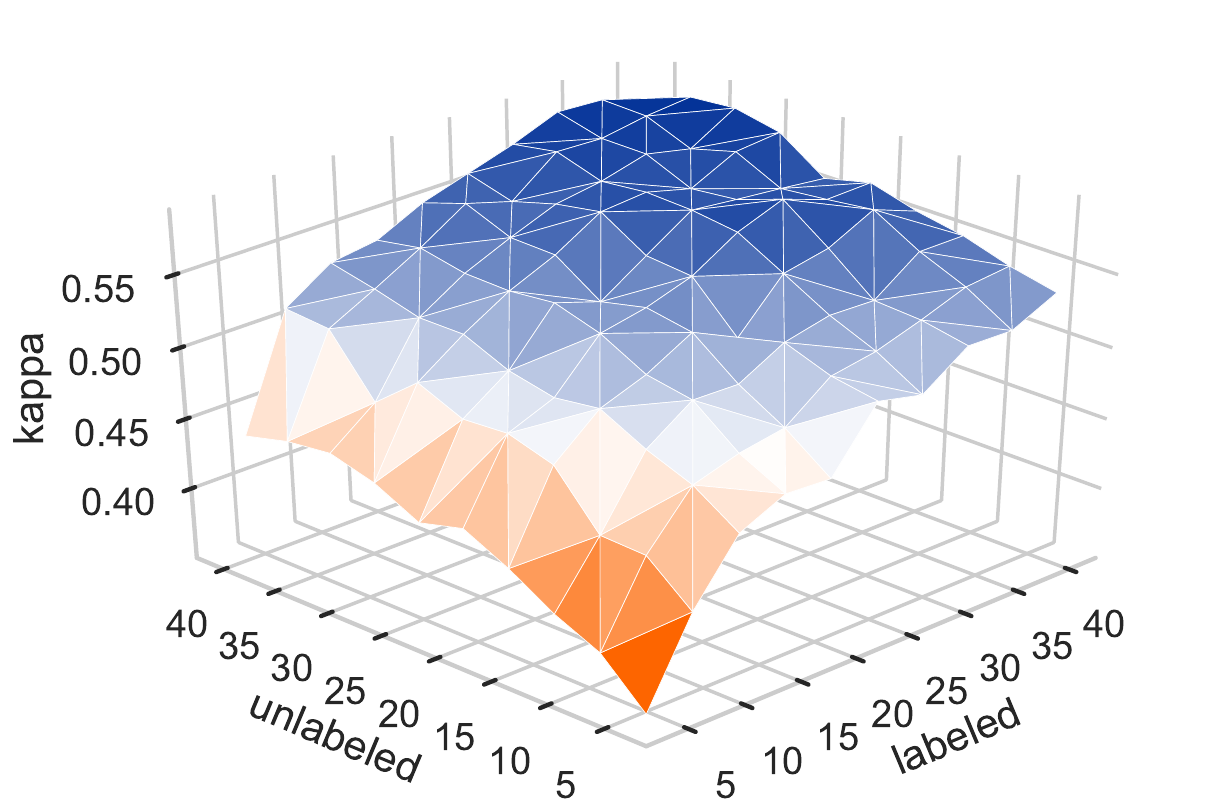}
\caption{Kappa.}
\label{fig:surfaceb}
\end{subfigure}
\begin{subfigure}{0.5\textwidth}
\centering
\includegraphics[height=4.75cm]{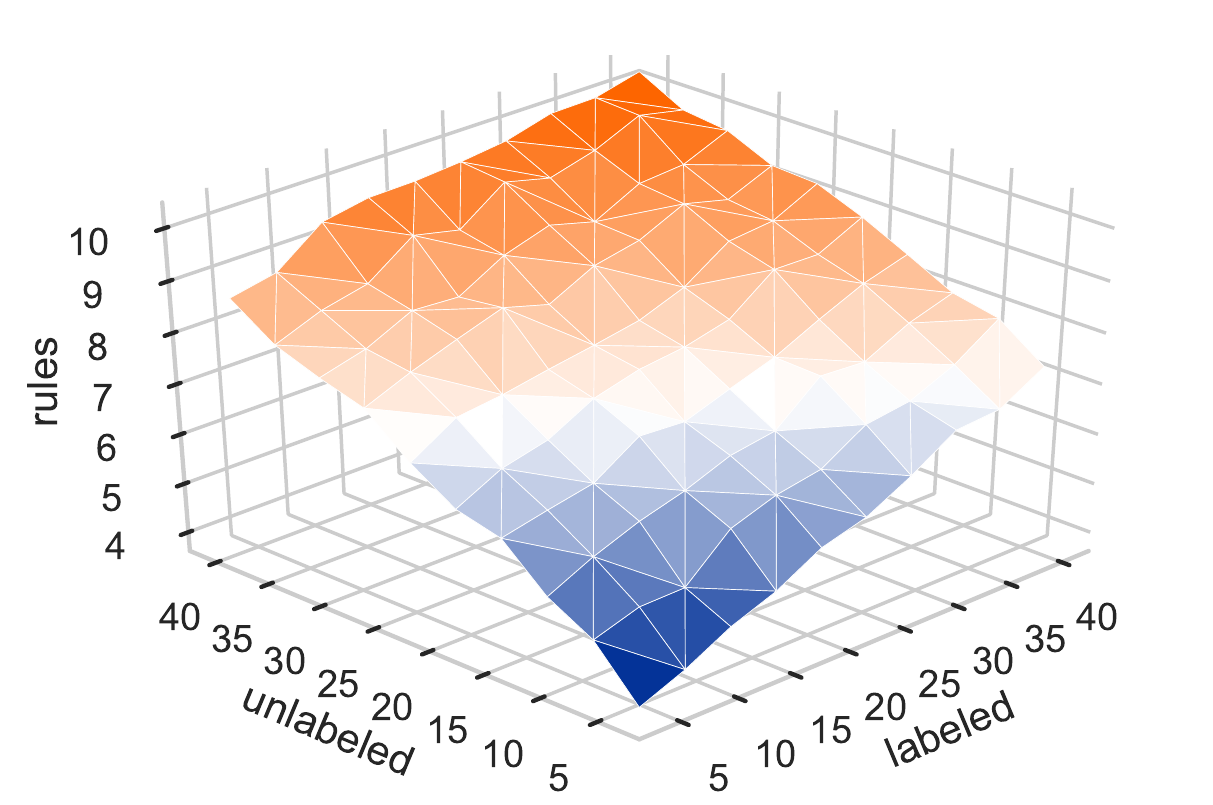}
\caption{Number of rules.}
\label{fig:surfacec}
\end{subfigure}
\begin{subfigure}{0.5\textwidth}
\centering
\includegraphics[height=4.75cm]{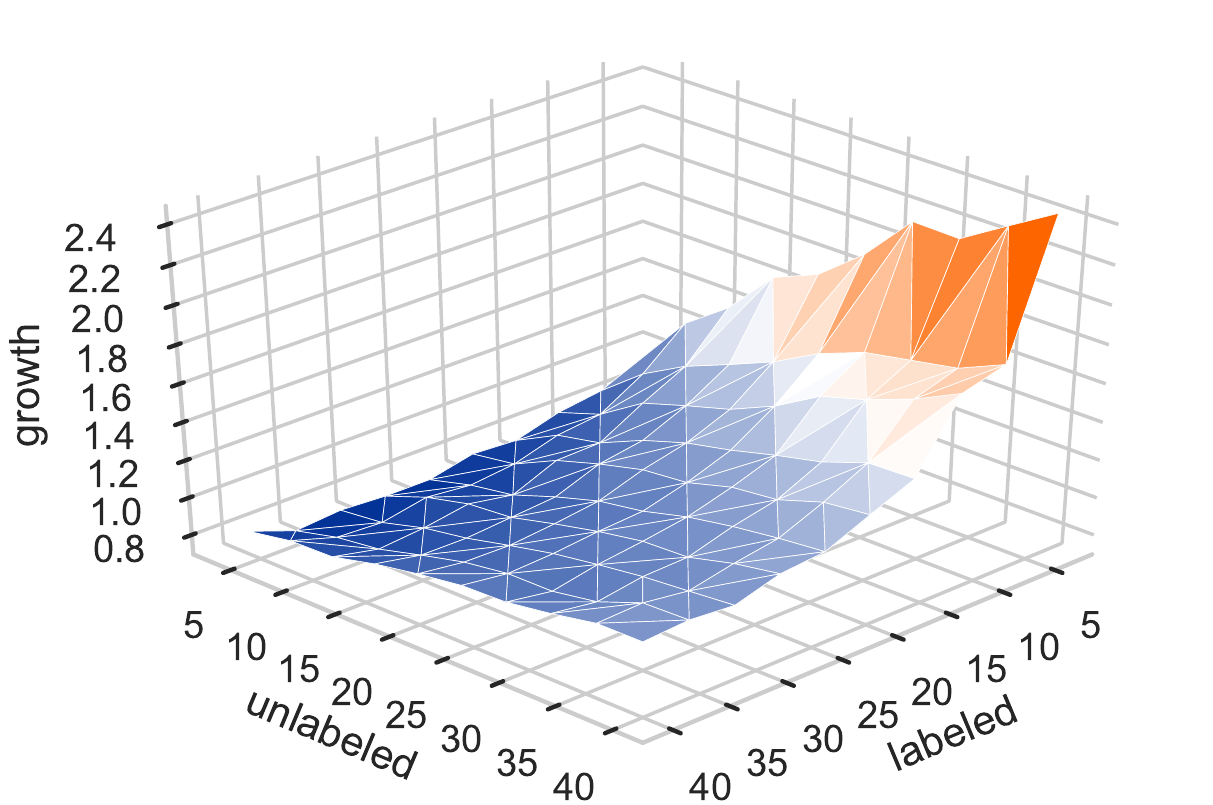}
\caption{Relative growth.}
\label{fig:surfaced}
\end{subfigure}
\begin{subfigure}{0.5\textwidth}
\centering
\includegraphics[height=4.75cm]{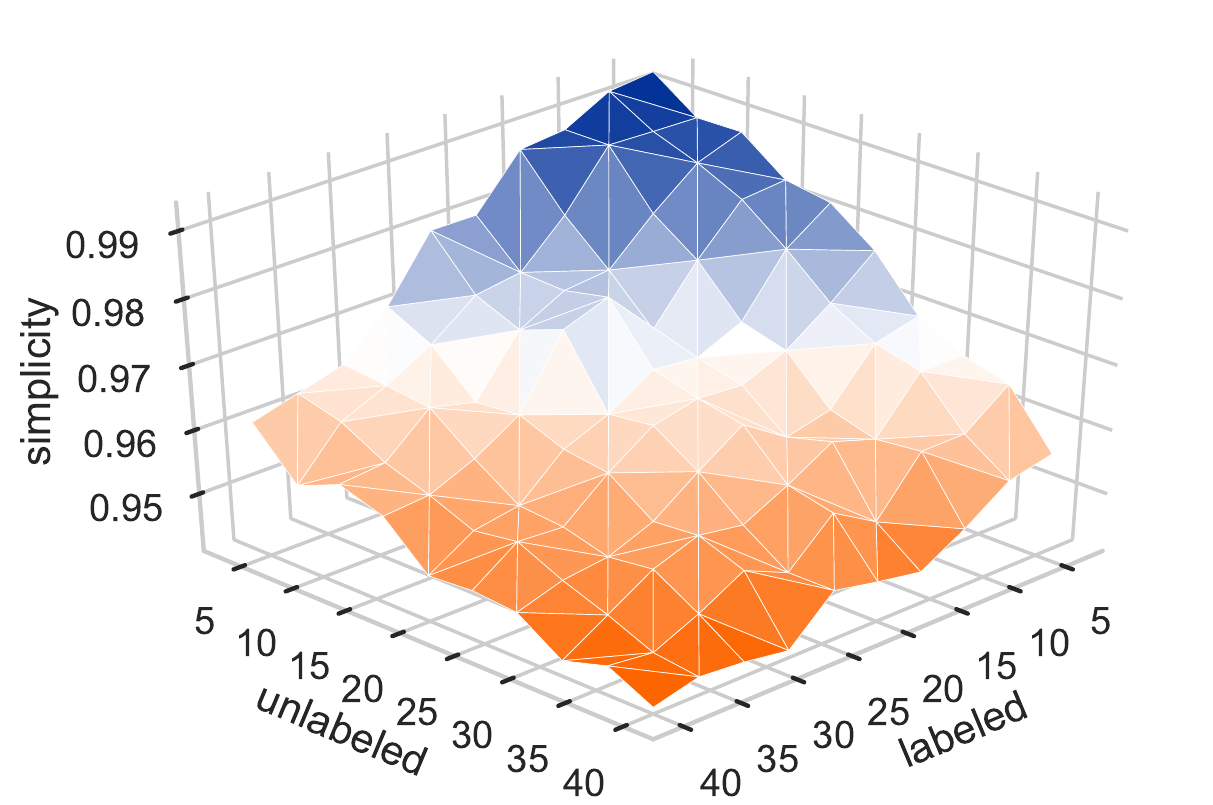}
\caption{Simplicity.}
\label{fig:surfacee}
\end{subfigure}
\begin{subfigure}{0.5\textwidth}
\centering
\includegraphics[height=4.75cm]{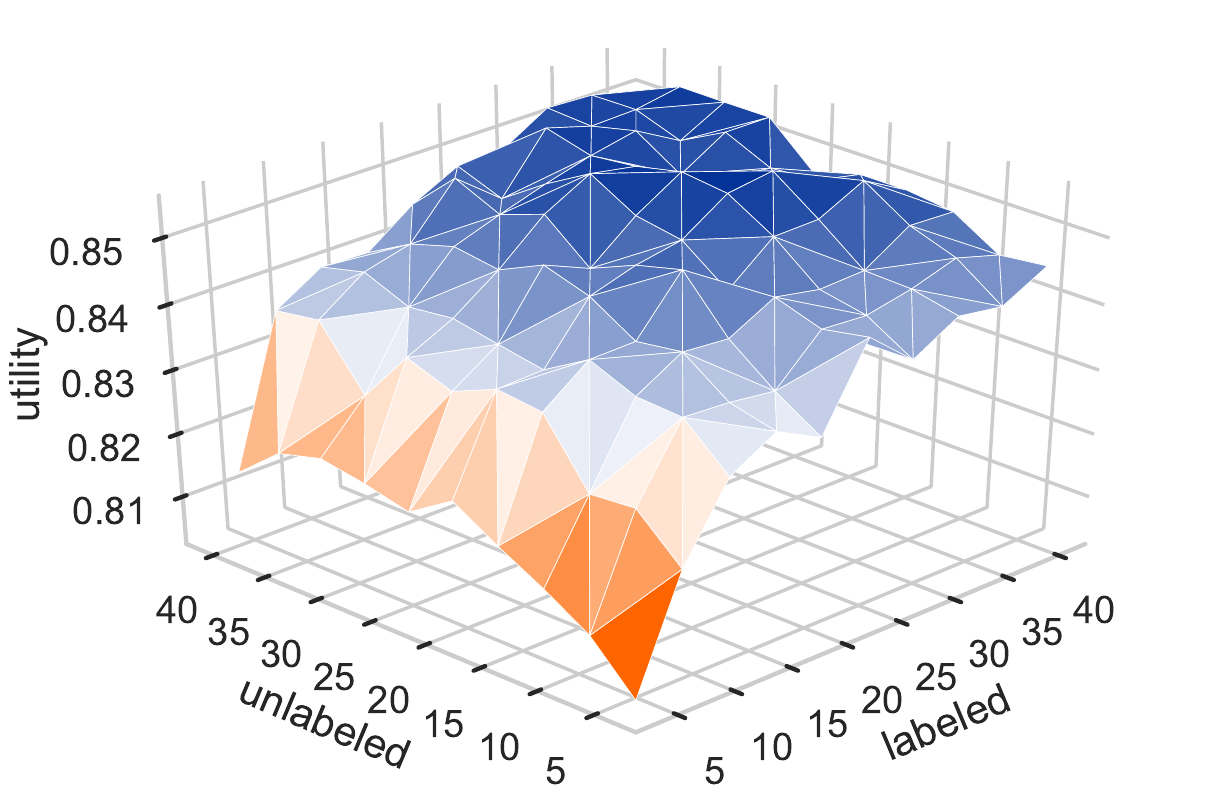}
\caption{Utility.}
\label{fig:surfacef}
\end{subfigure}
\caption{Performance of RF-RIP-RST when varying the number of labeled and unlabeled instances, for different measures. Axes $x$ and $y$ are expressed in percentage of instances taken for training from each subset. Sub-figures (d) and (e) are rotated for visualization purposes.}
\label{fig:surface}
\end{figure}

Due to the fact that we do not have truly unlabeled instances, we use the same datasets from the previous experiment.  First, a test set with 20\% of instances is kept aside for evaluation. Then, we divide the train set into two equally sized and disjoint subsets (each with 40\% of the total instances). Each subset is a source for labeled and unlabeled instances, respectively, from where we vary the amount of instances we use for training. Figure \ref{fig:surface} shows the surfaces resulting from the average of different measures over the 55 datasets. 

From the first two surfaces (Figures \ref{fig:surfacea} and \ref{fig:surfaceb}), it can be observed that the prediction rates (accuracy and kappa) have a pronounced increment when adding more labeled and unlabeled instances. The most dramatic change is observed when adding labeled data to a few unlabeled instances (5\%), which is an expected result as it tends to be a more supervised setting. However, when labeled instances are very limited (5\% of the dataset), adding unlabeled instances clearly increase the overall performance. In addition, even when more labeled data is available (40\%), an increase in performance is observed by adding more unlabeled data. This result confirms that our approach fulfills the main aim of SSC approaches. 

The number of rules (Figure \ref{fig:surfacec}) increases almost linearly with the number of training instances, either labeled or unlabeled. However, the relative growth (Figure \ref{fig:surfaced}) is more sensitive to adding unlabeled data when labeled data is very scarce, i.e. a bigger amount of unlabeled instances rapidly increases the structure and loses in interpretability, compared with the baseline white box. However, the base white boxes generally perform very poor when the labeled data is scarce. When the labeled data is not so scarce, then the growth is more robust to adding more unlabeled data. This means that even when a base white box can achieve good performance with some labeled data, adding unlabeled data does not generates too much growth in structure and can benefit the performance of the grey-box (Figures \ref{fig:surfacea} and \ref{fig:surfaceb}). 

The simplicity (Figure \ref{fig:surfacee}) shows the expected behavior: the best values of this measure are observed with the least number of instances and it decreases uniformly in both directions. This means that adding more unlabeled instances does not generate a greater number of extra rules compared to adding more labeled instances. This is a consequence of using amending procedures for adjusting the confidence of the unlabeled instances, thus avoiding that the white box learns from inconsistent instances. Finally, the utility surface (Fig. \ref{fig:surfacef}) summarizes all results reflecting the increase in the overall performance when adding both labeled and unlabeled instances.

\subsection{Comparing against State-of-the-Art Self-labeling Classifiers}
\label{sec:exp:cmp}

In this section we compare the predictive capability of SlGb against the four best self-labeling techniques reported in the review paper in \cite{triguero2015self}: Co-training using support vector machine \cite{hady2008co} (CT(SMO)), Tri-training using C45 decision tree \cite{zhou2005tri} (TT(C45)), Co-Bagging using C45 decision tree \cite{hady2008co} (CB(C45)) and Democratic Co-learning \cite{zhou2004democratic} (DCT). Since these algorithms are not inherently interpretable we focus our comparison on the prediction rates only. For this section, SlGb refers to the RF-PART-RST combination which exhibits the best results in kappa, as showed in Subsection \ref{sec:exp:white}.

Table \ref{table:sota} reports the mean and standard deviation of kappa coefficient for each classifier, taking into account the four studied ratios. The results reveal that SlGb has the highest mean for all ratios. In order to support this assertion, we compute the Friedman $p$-value per ratio. The test suggests rejecting the null hypotheses for all labeled ratios based on a confidence interval of 95\% (see Table \ref{table:friedman4}). This means that there is an indication that there exist significant differences between at least two algorithms in each comparison. 

\begin{table}[!ht]
\centering
\caption{Mean and standard deviation of kappa coefficient obtained by SlGb and four self-labeling methods from the state-of-the-art. The best performance is highlighted in bold.}
\begin{tabular}{|l|ll|ll|ll|ll|}
\thickhline
& \multicolumn{2}{c|}{10\%}          & \multicolumn{2}{c|}{20\%}           & \multicolumn{2}{c|}{30\%}         & \multicolumn{2}{c|}{40\%}         \\ 
& \multicolumn{1}{l}{mean} & stdev  & \multicolumn{1}{l}{mean} & stdev    & \multicolumn{1}{l}{mean} & stdev  & \multicolumn{1}{l}{mean} & stdev  \\ \hline
SlGb   & \textbf{0.56}  & \textbf{0.29}  & \textbf{0.61}    & \textbf{0.27}  & \textbf{0.62}    & \textbf{0.27}  & \textbf{0.62}    & \textbf{0.27}  \\
TT(C45) & 0.51 & 0.29  & 0.55 & 0.29  & 0.57 & 0.29  & 0.59 & 0.29  \\
CB(C45) & 0.51 & 0.29  & 0.55 & 0.29  & 0.57 & 0.29  & 0.56 & 0.28  \\
DCT     & 0.49 & 0.32  & 0.54 & 0.30  & 0.58 & 0.28  & 0.59 & 0.28  \\
CT(SMO) & 0.48 & 0.31  & 0.55 & 0.29  & 0.58 & 0.29  & 0.60 & 0.29  \\ \thickhline
\end{tabular}
\label{table:sota}
\end{table}

The next step is focused on determining whether the superiority of the SlGb classifier is responsible for the significant difference reported by the Friedman test. Similar to previous sections we use the Wilcoxon signed rank test and the Holm post-hoc procedure for computing the corrected $p$-values associated with each pairwise comparison. Each section of the Table \ref{table:wilcoxon4} represents a ratio of labeled instances. The null hypothesis states that there is no significant difference between the performance of each pair of algorithms, taking SlGb as the control one.

From the statistical tests we can draw the following conclusions. First, there is no doubt about the superiority of the SlGb classifier when tested with datasets with ratios of 10\% and 20\% of labeled instances, as all the null hypotheses were rejected. This result, in combination with the first place in the Friedman ranking, demonstrates that our algorithm significantly outperforms the other four algorithms in these settings. In the case of datasets comprising 30\% and 40\% of labeled instances, the results show that SlGb is the best-performing classifier, but with no significant differences observed between the pairs SlGb vs. DCT (for 30\%), and SlGb vs. CT(SMO) (for both ratios), as these null hypotheses could not be rejected. However, DCT and CT(SMO) cannot be considered transparent due to their complex structure involving support vector machines and collaboration between base classifiers. Although our main goal was not to outperform the SSC methods in terms of classification rates, the analysis reported above supports our claim that we obtain a favorable balance between performance and interpretability by using the SlGb approach for solving SSC problems.

\section{Conclusions}
\label{sec:conclusions}

In this paper, we report on an extended experimental study to determine the suitability of SlGb classifier for semi-supervised classification problems where interpretability is a requirement. We explore two different amending procedures for weighting the instances coming from the self-labeling process. Such procedures aim at preventing the effect of misclassifications to propagate across the whole model. The experiments have shown that using Random Forests as the base black box for the self-labeling process is the best choice in terms of prediction rates. The choice of a white box and amending does not significantly affect the prediction rates but it is relevant for the size of the structure. 

Three measures based on the number of rules were proposed for estimating the relative growth, simplicity, and utility of the SlGb. SlGb produces simpler models when using decision lists instead of a C4.5 decision tree as surrogate white boxes, even when no amending is performed. However, the amending procedures help further increase the simplicity (and therefore transparency) without affecting the prediction rates by giving more importance to confident instances in the self-labeling. Especially RST based amending looks more promising since it does not need the black-box base classifier to provide calibrated probabilities. Furthermore, RST based amending could be a good choice for a given case study where the uncertainty coming from inconsistency is high, even on the available labeled data. Therefore, we strongly advise the use of random forests as a base black box and RST for amending the self-labeling, while the choice of white box is more flexible to the desired interpretability, either a decision tree with rules or a decision list. Although, the best trade-off between accuracy and interpretability (utility) is reached when using the RF-RIP-RST combination. 

The study varying the number of unlabeled instances and labeled instances together shows that even when the number of labeled instances is not that scarce, the SlGb is able to leverage unlabeled instances for increasing the performance. Another conclusion is that adding unlabeled instances does not make the interpretability worse compared to adding more labeled instances. This evidences that the amending procedure (in this case RST-based amending) avoids that the SlGb generates more rules from inconsistent instances. Finally, the experimental comparison shows that our SlGb method outperforms the state-of-the-art self-labeling approaches, yet being far more simple in structure than these techniques.

\bibliographystyle{unsrt} 
\bibliography{references}

\clearpage
\section{Appendix: Benchmark Datasets and Detailed Results of Statistical Tests}

\begin{center}
\begin{longtable}{|lcccccccc|}
\caption{Characterization of the datasets used in experiments in Section \ref{sec:exp}. The imbalance is computed as the ratio of the number of instances between the majority and the minority class of the dataset, NA means a ratio smaller than two.}\\

\hline
\multicolumn{1}{|l}{Dataset} & \multicolumn{1}{c}{Instances} & \multicolumn{1}{c}{10\%} & \multicolumn{1}{c}{20\%} & \multicolumn{1}{c}{30\%} & \multicolumn{1}{c}{40\%} & \multicolumn{1}{c}{Features} & \multicolumn{1}{c}{Classes} & \multicolumn{1}{c|}{Imbalance}  \\
\hline
\endfirsthead
\multicolumn{9}{c}%
{\tablename\ \thetable\ -- \textit{Continued from previous page}} \\
\hline
\multicolumn{1}{|l}{Dataset} & \multicolumn{1}{c}{Instances} & \multicolumn{1}{c}{10\%} & \multicolumn{1}{c}{20\%} & \multicolumn{1}{c}{30\%} & \multicolumn{1}{c}{40\%} & \multicolumn{1}{c}{Features} & \multicolumn{1}{c}{Classes} & \multicolumn{1}{c|}{Imbalance}  \\
\hline
\endhead
\hline 
\multicolumn{9}{r}{\textit{Continued on next page}} \\
\endfoot
\hline
\endlastfoot
\label{table:datasets}
abalone          & 4,174       & 417  & 835  & 1,252 & 1,670 & 8          & 28       & 689.00 \\
appendicitis     & 106        & 11   & 21   & 32   & 42   & 7          & 2        & 4.04 \\
australian       & 690        & 69   & 138  & 207  & 276  & 14         & 2        & NA \\
autos            & 205        & 21   & 41   & 62   & 82   & 25         & 6        & 22.33 \\
banana           & 5,300       & 530  & 1,060 & 1,590 & 2,120 & 2          & 2        & NA \\
breast           & 286        & 29   & 57   & 86   & 114  & 9          & 2        & 2.36 \\
bupa             & 345        & 35   & 69   & 104  & 138  & 6          & 2        & NA \\
chess            & 3,196       & 320  & 639  & 959  & 1,278 & 36         & 2        & NA \\
cleveland        & 297        & 30   & 59   & 89   & 119  & 13         & 5        & 12.61 \\
coil2000         & 9,822       & 982  & 1,964 & 2,947 & 3,929 & 85         & 2        & 15.76 \\
contraceptive    & 1,473       & 147  & 295  & 442  & 589  & 9          & 3        & NA \\
crx              & 125        & 13   & 25   & 38   & 50   & 15         & 2        & NA \\
dermatology      & 366        & 37   & 73   & 110  & 146  & 33         & 6        & 5.6 \\
ecoli            & 336        & 34   & 67   & 101  & 134  & 7          & 8        & 71.5\\
flare-solar      & 1,066       & 107  & 213  & 320  & 426  & 9          & 6        & 7.69 \\
german           & 1,000       & 100  & 200  & 300  & 400  & 20         & 2        & 2.33 \\
glass            & 214        & 21   & 43   & 64   & 86   & 9          & 7        & 8.44 \\
haberman         & 306        & 31   & 61   & 92   & 122  & 3          & 2        & 2.77 \\
heart            & 270        & 27   & 54   & 81   & 108  & 13         & 2        & NA \\
hepatitis        & 155        & 16   & 31   & 47   & 62   & 19         & 2        & 3.84 \\
housevotes       & 435        & 44   & 87   & 131  & 174  & 16         & 2        & NA \\
iris             & 150        & 15   & 30   & 45   & 60   & 4          & 3        & NA \\
led7digit        & 500        & 50   & 100  & 150  & 200  & 7          & 10       & NA \\
lymphography     & 148        & 15   & 30   & 44   & 59   & 18         & 4        & 40.5 \\
magic            & 19,020      & 1,902 & 3,804 & 5,706 & 7,608 & 10         & 2        & NA \\
mammographic     & 961        & 96   & 192  & 288  & 384  & 5          & 2        & NA \\
marketing        & 8,993       & 899  & 1,799 & 2,698 & 3,597 & 13         & 9        & 2.48 \\
monks            & 432        & 43   & 86   & 130  & 173  & 6          & 2        & NA \\
movement\_libras & 360        & 36   & 72   & 108  & 144  & 90         & 15       & NA \\
mushroom         & 8,124       & 812  & 1,625 & 2,437 & 3,250 & 22         & 2        & NA \\
nursery          & 12,690      & 1,269 & 2,538 & 3,807 & 5,076 & 8          & 5        & 2,160.00 \\
pageblocks       & 5,472       & 547  & 1,094 & 1,642 & 2,189 & 10         & 5        & 175.46 \\
penbased         & 10,992      & 1,099 & 2,198 & 3,298 & 4,397 & 16         & 10       & NA \\
phoneme          & 5,404       & 540  & 1,081 & 1,621 & 2,162 & 5          & 2        & 2.4 \\
pima             & 768        & 77   & 154  & 230  & 307  & 8          & 2        & NA \\
ring             & 7,400       & 740  & 1,480 & 2,220 & 2,960 & 20         & 2        & NA \\
saheart          & 462        & 46   & 92   & 139  & 185  & 9          & 2        & NA \\
satimage         & 6,435       & 644  & 1,287 & 1,931 & 2,574 & 36         & 7        & 2.44 \\
segment          & 2,310       & 231  & 462  & 693  & 924  & 19         & 7        & NA \\
sonar            & 208        & 21   & 42   & 62   & 83   & 60         & 2        & NA \\
spambase         & 4,597       & 460  & 919  & 1,379 & 1,839 & 55         & 2        & NA \\
spectheart       & 267        & 27   & 53   & 80   & 107  & 44         & 2        & 3.85 \\
splice           & 3,190       & 319  & 638  & 957  & 1,276 & 60         & 3        & 2.15 \\
tae              & 151        & 15   & 30   & 45   & 60   & 5          & 3        & NA \\
texture          & 5,500       & 550  & 1,100 & 1,650 & 2,200 & 40         & 11       & NA \\
tic-tac-toe      & 958        & 96   & 192  & 287  & 383  & 9          & 2        & NA \\
thyroid          & 7,200       & 720  & 1,440 & 2,160 & 2,880 & 21         & 3        & 40.15 \\
titanic          & 2,201       & 220  & 440  & 660  & 880  & 3          & 2        & 2.09 \\
twonorm          & 7,400       & 740  & 1,480 & 2,220 & 2,960 & 20         & 2        & NA \\
vehicle          & 846        & 85   & 169  & 254  & 338  & 18         & 4        & NA \\
vowel            & 990        & 99   & 198  & 297  & 396  & 13         & 11       & NA \\
wine             & 178        & 18   & 36   & 53   & 71   & 13         & 3        & NA  \\
wisconsin        & 683        & 68   & 137  & 205  & 273  & 9          & 2        & NA \\
yeast            & 1,484       & 148  & 297  & 445  & 594  & 8          & 10       & 92.6 \\
zoo              & 101        & 10   & 20   & 30   & 40   & 17         & 7        & 10.25 \\
\hline
\end{longtable}
\end{center}

\begin{table}[!ht]
\centering
\caption{Friedman $p$-values for all ratios when testing different black-box base classifiers. The prediction rates are measured using kappa coefficient. There are significant differences among all the configurations compared.}
\label{table:friedman1}
\begin{tabular}{|c|c|c|}
\thickhline
Ratio & Friedman $p$-value & $H_0$ \\ \thickhline
10\% & 2.10E-08 & Rejected \\
20\% & 8.91E-13 & Rejected \\
30\% & 9.87E-06 & Rejected \\
40\% & 5.35E-04 & Rejected \\ \thickhline
\end{tabular}
\end{table}

\begin{table}[!ht]
\centering
\caption{Wilcoxon $p$-values and Holm's post-hoc correction when comparing different black-boxes configurations. The test supports the superiority of RF as black-box base classifier when comparing prediction rates.}
\label{table:wilcoxon1}
\resizebox{\textwidth}{!} {
\begin{tabular}{|c|c|cccc|c|}
\thickhline
Labeled ratio & Pair of configurations & Wilcoxon $p$-value & $R^-$ & $R^+$ & Holm &  $H_0$ \\
\thickhline
\multirow{6}{*}{10\%} & RF-PART - MLP-PART & 2.08E-05 & 13 & 39 & 1.25E-04 & Rejected \\
                      & RF-PART - SVM-PART & 2.77E-04 & 16 & 39 & 1.38E-03 & Rejected \\
                      & RF-C45 - SVM-C45 & 6.91E-04 & 16 & 39 & 2.76E-03 & Rejected \\
                      & RF-RIP - MLP-RIP & 9.63E-04 & 15 & 40 & 2.89E-03 & Rejected \\
                      & RF-C45 - MLP-C45 & 1.6E-03 & 15 & 39 & 3.2E-03 & Rejected \\
                      & RF-RIP - SVM-RIP & 5.84E-03 & 19 & 36 & 5.84E-03 & Rejected \\
\thickhline
\multirow{6}{*}{20\%} & RF-C45 - MLP-C45 & 5.58E-06 & 9 & 45 & 3.35E-05 & Rejected \\
                      & RF-PART - SVM-PART & 2.1E-05 & 15 & 39 & 1.05E-04 & Rejected \\
                      & RF-PART - MLP-PART & 4.56E-05 & 13 & 41 & 1.83E-04 & Rejected \\
                      & RF-RIP - SVM-RIP & 1.83E-04 & 17 & 37 & 5.5E-04 & Rejected \\
                      & RF-C45 - SVM-C45 & 3.04E-04 & 15 & 39 & 6.08E-04 & Rejected \\
                      & RF-RIP - MLP-RIP & 3.56E-03 & 18 & 36 & 3.56E-03 & Rejected \\
\thickhline
\multirow{6}{*}{30\%} & RF-RIP - MLP-RIP & 1.18E-03 & 15 & 38 & 7.06E-03 & Rejected \\
                      & RF-C45 - MLP-C45 & 1.19E-03 & 16 & 38 & 7.06E-03 & Rejected \\
                      & RF-C45 - SVM-C45 & 1.75E-03 & 16 & 38 & 7.06E-03 & Rejected \\
                      & RF-RIP - SVM-RIP & 2.93E-03 & 18 & 36 & 8.79E-03 & Rejected \\
                      & RF-PART - MLP-PART & 5.64E-03 & 20 & 34 & 1.13E-02 & Rejected \\
                      & RF-PART - SVM-PART & 7.7E-03 & 20 & 34 & 1.13E-02 & Rejected \\
\thickhline
\multirow{6}{*}{40\%} & RF-RIP - SVM-RIP & 1.55E-03 & 16 & 38 & 9.33E-03 & Rejected \\
                      & RF-PART - MLP-PART & 6.26E-03 & 19 & 35 & 3.13E-02 & Rejected \\
                      & RF-C45 - MLP-C45 & 8.75E-03 & 20 & 34 & 3.5E-02 & Rejected \\
                      & RF-PART - SVM-PART & 1.43E-02 & 19 & 35 & 4.29E-02 & Rejected \\
                      & RF-C45 - SVM-C45 & 2.28E-02 & 22 & 32 & 4.55E-02 & Rejected \\
                      & RF-RIP - MLP-RIP & 7.68E-02 & 23 & 31 & 7.68E-02 & Not Rejected \\
\thickhline
\end{tabular}
}
\end{table}

\begin{table}[!ht]
\centering
\caption{Friedman $p$-values for all ratios when testing the prediction performance (kappa) for different white-box and amending configurations. There are statistical differences in the prediction rates in at least one pair of the configurations compared.}
\label{table:friedman2}
\begin{tabular}{|c|c|c|}
\thickhline
Ratio & Friedman $p$-value & $H_0$ \\ \thickhline
10\% & 4.02E-07 & Rejected \\
20\% & 9.27E-07 & Rejected \\
30\% & 1.67E-03 & Rejected \\
40\% & 7.35E-05 & Rejected \\ \thickhline
\end{tabular}
\end{table}

\begin{table}[!ht]
\centering
\caption{Wilcoxon's $p$-values and Holm's post-hoc correction when comparing different white-box and amending configurations, for 10\% and 20\% ratio. Per ratio, first subsection compares using different amending procedures while fixing the white box and the second subsection fixes the amending for comparing the influence of white boxes. The vast majority of null hypothesis cannot be rejected, indicating that amending or white-box alternatives do not strongly influence the prediction rates.}
\label{table:wilcoxon2a}
\resizebox{\textwidth}{!} {
\begin{tabular}{|c|c|cccc|c|}
\hline
Labeled ratio & Pair of configurations & Wilcoxon's $p$-value & $R^-$ & $R^+$ & Holm's $p$-value &  $H_0$ \\
\hline
\multirow{15}{*}{10\%}    & RF-RIP-RST - RF-RIP-NONE & 2.57E-03 & 40 & 12 & 1.54E-02 & Rejected \\
                         & RF-C45-RST - RF-C45-NONE & 3.4E-02 & 36 & 17 & 0.170 & Not Rejected \\
                         & RF-RIP-CONF - RF-RIP-NONE & 4.23E-02 & 33 & 19 & 0.170 & Not Rejected \\
                         & RF-PART-RST - RF-PART-NONE & 5.24E-02 & 31 & 21 & 0.170 & Not Rejected \\
                         & RF-C45-CONF - RF-C45-NONE & 0.224 & 30 & 23 & 0.447 & Not Rejected \\
                         & RF-PART-CONF - RF-PART-NONE & 0.344 & 27 & 25 & 0.447 & Not Rejected \\
                         \cline{2-7}
                         & RF-RIP-CONF - RF-PART-CONF & 7.09E-04 & 35 & 18 & 6.38E-03 & Rejected \\
                         & RF-RIP-RST - RF-PART-RST & 1.44E-02 & 35 & 18 & 0.115 & Not Rejected \\
                         & RF-RIP-NONE - RF-PART-NONE & 1.97E-02 & 34 & 19 & 0.138 & Not Rejected \\
                         & RF-RIP-RST - RF-C45-RST & 2.01E-02 & 33 & 20 & 0.138 & Not Rejected \\
                         & RF-RIP-CONF - RF-C45-CONF & 3.25E-02 & 33 & 20 & 0.163 & Not Rejected \\
                         & RF-PART-CONF - RF-C45-CONF & 9.52E-02 & 18 & 33 & 0.381 & Not Rejected \\
                         & RF-PART-NONE - RF-C45-NONE & 0.166 & 23 & 29 & 0.499 & Not Rejected \\
                         & RF-PART-RST - RF-C45-RST & 0.188 & 20 & 30 & 0.499 & Not Rejected \\
                         & RF-RIP-NONE - RF-C45-NONE & 0.510 & 31 & 23 & 0.510 & Not Rejected \\
\hline
\multirow{15}{*}{20\%} & RF-RIP-RST - RF-RIP-NONE & 8.04E-05 & 38 & 14 & 4.82E-04 & Rejected \\
                     & RF-C45-CONF - RF-C45-NONE & 4.89E-03 & 36 & 14 & 2.45E-02 & Rejected \\
                     & RF-PART-RST - RF-PART-NONE & 4.92E-02 & 33 & 19 & 0.197 & Not Rejected \\
                     & RF-PART-CONF - RF-PART-NONE & 5.23E-02 & 33 & 18 & 0.197 & Not Rejected \\
                     & RF-C45-RST - RF-C45-NONE & 5.82E-02 & 34 & 18 & 0.197 & Not Rejected \\
                     & RF-RIP-CONF - RF-RIP-NONE & 0.169 & 31 & 21 & 0.197 & Not Rejected \\
                     \cline{2-7}
                     & RF-RIP-RST - RF-PART-RST & 1.6E-03 & 39 & 14 & 1.44E-02 & Rejected \\
                     & RF-RIP-RST - RF-C45-RST & 6.12E-03 & 36 & 16 & 4.9E-02 & Rejected \\
                     & RF-RIP-NONE - RF-PART-NONE & 2.78E-02 & 36 & 17 & 0.195 & Not Rejected \\
                     & RF-RIP-NONE - RF-C45-NONE & 4.53E-02 & 36 & 18 & 0.272 & Not Rejected \\
                     & RF-RIP-CONF - RF-C45-CONF & 0.169 & 30 & 22 & 0.854 & Not Rejected \\
                     & RF-RIP-CONF - RF-PART-CONF & 0.259 & 28 & 25 & 1.000 & Not Rejected \\
                     & RF-PART-RST - RF-C45-RST & 0.769 & 26 & 23 & 1.000 & Not Rejected \\
                     & RF-PART-NONE - RF-C45-NONE & 0.827 & 27 & 25 & 1.000 & Not Rejected \\
                     & RF-PART-CONF - RF-C45-CONF & 0.889 & 27 & 23 & 1.000 & Not Rejected \\
\hline
\end{tabular}
}
\end{table}

\begin{table}[!ht]
\centering
\caption{Wilcoxon's $p$-values and Holm's post-hoc correction when comparing different white-box and amending configurations, for 30\% and 40\% ratio. Per ratio, first subsection compares using different amending procedures while fixing the white box and the second subsection fixes the amending for comparing the influence of white boxes. The vast majority of null hypothesis cannot be rejected, indicating that amending or white-box alternatives do not strongly influence the prediction rates.}
\label{table:wilcoxon2b}
\resizebox{\textwidth}{!} {
\begin{tabular}{|c|c|cccc|c|}
\hline
Labeled ratio & Pair of configurations & Wilcoxon's $p$-value & $R^-$ & $R^+$ & Holm's $p$-value &  $H_0$ \\
\hline
\multirow{15}{*}{30\%}  & RF-RIP-RST - RF-RIP-NONE & 2.18E-04 & 37 & 15 & 1.31E-03 & Rejected \\
                     & RF-C45-RST - RF-C45-NONE & 8.45E-03 & 37 & 16 & 4.22E-02 & Rejected \\
                     & RF-RIP-CONF - RF-RIP-NONE & 4.14E-02 & 32 & 20 & 0.165 & Not Rejected \\
                     & RF-C45-CONF - RF-C45-NONE & 0.193 & 28 & 24 & 0.578 & Not Rejected \\
                     & RF-PART-CONF - RF-PART-NONE & 0.362 & 28 & 24 & 0.725 & Not Rejected \\
                     & RF-PART-RST - RF-PART-NONE & 0.388 & 28 & 25 & 0.725 & Not Rejected \\
                     \cline{2-7}
                     & RF-RIP-RST - RF-PART-RST & 1.68E-03 & 37 & 15 & 1.51E-02 & Rejected \\
                     & RF-RIP-RST - RF-C45-RST & 7.03E-03 & 35 & 17 & 5.62E-02 & Not Rejected \\
                     & RF-RIP-CONF - RF-C45-CONF & 0.147 & 30 & 24 & 1.000 & Not Rejected \\
                     & RF-PART-RST - RF-C45-RST & 0.259 & 22 & 27 & 1.000 & Not Rejected \\
                     & RF-RIP-CONF - RF-PART-CONF & 0.324 & 30 & 24 & 1.000 & Not Rejected \\
                     & RF-RIP-NONE - RF-PART-NONE & 0.355 & 31 & 22 & 1.000 & Not Rejected \\
                     & RF-RIP-NONE - RF-C45-NONE & 0.498 & 28 & 25 & 1.000 & Not Rejected \\
                     & RF-PART-NONE - RF-C45-NONE & 0.555 & 22 & 29 & 1.000 & Not Rejected \\
                     & RF-PART-CONF - RF-C45-CONF & 0.573 & 23 & 25 & 1.000 & Not Rejected \\
\hline
\multirow{15}{*}{40\%}  & RF-RIP-RST - RF-RIP-NONE & 1.35E-05 & 40 & 13 & 8.13E-05 & Rejected \\
                     & RF-RIP-CONF - RF-RIP-NONE & 7.03E-03 & 31 & 21 & 3.51E-02 & Rejected \\
                     & RF-PART-RST - RF-PART-NONE & 0.136 & 30 & 23 & 0.543 & Not Rejected \\
                     & RF-PART-CONF - RF-PART-NONE & 0.579 & 28 & 24 & 1.000 & Not Rejected \\
                     & RF-C45-RST - RF-C45-NONE & 0.662 & 26 & 26 & 1.000 & Not Rejected \\
                     & RF-C45-CONF - RF-C45-NONE & 0.761 & 26 & 24 & 1.000 & Not Rejected \\
                     \cline{2-7}
                      & RF-RIP-RST - RF-C45-RST & 2.9E-03 & 39 & 13 & 2.61E-02 & Rejected \\
                     & RF-RIP-CONF - RF-PART-CONF & 2.13E-02 & 36 & 18 & 0.170 & Not Rejected \\
                     & RF-RIP-CONF - RF-C45-CONF & 3.25E-02 & 33 & 20 & 0.228 & Not Rejected \\
                     & RF-RIP-RST - RF-PART-RST & 3.96E-02 & 33 & 20 & 0.237 & Not Rejected \\
                     & RF-RIP-NONE - RF-PART-NONE & 0.254 & 33 & 21 & 1.000 & Not Rejected \\
                     & RF-PART-NONE - RF-C45-NONE & 0.267 & 23 & 30 & 1.000 & Not Rejected \\
                     & RF-PART-CONF - RF-C45-CONF & 0.606 & 26 & 24 & 1.000 & Not Rejected \\
                     & RF-RIP-NONE - RF-C45-NONE & 0.806 & 29 & 23 & 1.000 & Not Rejected \\
                     & RF-PART-RST - RF-C45-RST & 0.858 & 26 & 24 & 1.000 & Not Rejected \\
\hline
\end{tabular}
}
\end{table}

\begin{table}[!ht]
\centering
\caption{Friedman $p$-values for all ratios when comparing the interpretability in terms of simplicity, for different white-box and amending configurations. There are significant differences among all the configurations compared, where RF-RIP-RST exhibits the highest mean for all ratios (see Table \ref{table:performance:interp}).}
\label{table:friedman3}
\begin{tabular}{|c|c|c|}
\thickhline
Ratio & Friedman $p$-value & $H_0$ \\ \thickhline
10\% & 3.14E-73 & Rejected \\
20\% & 3.02E-75 & Rejected \\
30\% & 3.96E-72 & Rejected \\
40\% & 2.41E-71 & Rejected \\ \thickhline
\end{tabular}
\end{table}

\begin{table}[!ht]
\centering
\caption{Wilcoxon $p$-values and Holm's post-hoc correction when comparing different white-box and amending configurations against the highest mean simplicity combination: RF-RIP-RST. All null hypothesis can be safely rejected, showing statistically significant superiority in terms of simplicity.}
\label{table:wilcoxon3}
\resizebox{\textwidth}{!} {
\begin{tabular}{|c|c|cccc|c|}
\thickhline
Labeled ratio & Pair of configurations & Wilcoxon $p$-value & $R^-$ & $R^+$ & Holm &  $H_0$ \\
\thickhline
\multirow{8}{*}{10\%}  & RF-RIP-RST - RF-C45-NONE & 1.11E-10 & 0 & 55 & 8.86E-10 & Rejected \\
                     & RF-RIP-RST - RF-PART-NONE & 1.17E-10 & 1 & 54 & 8.86E-10 & Rejected \\
                     & RF-RIP-RST - RF-C45-CONF & 1.63E-10 & 0 & 54 & 9.75E-10 & Rejected \\
                     & RF-RIP-RST - RF-C45-RST & 1.63E-10 & 0 & 54 & 9.75E-10 & Rejected \\
                     & RF-RIP-RST - RF-RIP-NONE & 2.39E-10 & 0 & 53 & 9.75E-10 & Rejected \\
                     & RF-RIP-RST - RF-PART-CONF & 3.97E-10 & 2 & 52 & 1.19E-09 & Rejected \\
                     & RF-RIP-RST - RF-PART-RST & 4.08E-09 & 3 & 51 & 8.17E-09 & Rejected \\
                     & RF-RIP-RST - RF-RIP-CONF & 2.18E-07 & 5 & 46 & 2.18E-07 & Rejected \\
\thickhline
\multirow{8}{*}{20\%}  & RF-RIP-RST - RF-C45-NONE & 1.11E-10 & 0 & 55 & 8.86E-10 & Rejected \\
                     & RF-RIP-RST - RF-PART-NONE & 1.11E-10 & 0 & 55 & 8.86E-10 & Rejected \\
                     & RF-RIP-RST - RF-C45-CONF & 1.11E-10 & 0 & 55 & 8.86E-10 & Rejected \\
                     & RF-RIP-RST - RF-PART-CONF & 1.11E-10 & 0 & 55 & 8.86E-10 & Rejected \\
                     & RF-RIP-RST - RF-C45-RST & 1.63E-10 & 0 & 54 & 8.86E-10 & Rejected \\
                     & RF-RIP-RST - RF-RIP-NONE & 2.39E-10 & 0 & 53 & 8.86E-10 & Rejected \\
                     & RF-RIP-RST - RF-RIP-CONF & 8.76E-10 & 1 & 52 & 1.75E-09 & Rejected \\
                     & RF-RIP-RST - RF-PART-RST & 3.95E-08 & 3 & 50 & 3.95E-08 & Rejected \\
\thickhline
\multirow{8}{*}{30\%}  & RF-RIP-RST - RF-C45-NONE & 1.11E-10 & 0 & 55 & 8.86E-10 & Rejected \\
                     & RF-RIP-RST - RF-C45-CONF & 1.11E-10 & 0 & 55 & 8.86E-10 & Rejected \\
                     & RF-RIP-RST - RF-PART-NONE & 1.24E-10 & 1 & 54 & 8.86E-10 & Rejected \\
                     & RF-RIP-RST - RF-C45-RST & 1.24E-10 & 1 & 54 & 8.86E-10 & Rejected \\
                     & RF-RIP-RST - RF-PART-CONF & 1.31E-10 & 2 & 53 & 8.86E-10 & Rejected \\
                     & RF-RIP-RST - RF-RIP-NONE & 2.6E-10 & 1 & 52 & 8.86E-10 & Rejected \\
                     & RF-RIP-RST - RF-RIP-CONF & 8.03E-10 & 1 & 49 & 1.61E-09 & Rejected \\
                     & RF-RIP-RST - RF-PART-RST & 1.18E-09 & 3 & 51 & 1.61E-09 & Rejected \\
\thickhline
\multirow{8}{*}{40\%}  & RF-RIP-RST - RF-C45-NONE & 1.11E-10 & 0 & 55 & 8.86E-10 & Rejected \\
                     & RF-RIP-RST - RF-C45-CONF & 1.11E-10 & 0 & 55 & 8.86E-10 & Rejected \\
                     & RF-RIP-RST - RF-PART-NONE & 1.24E-10 & 1 & 54 & 8.86E-10 & Rejected \\
                     & RF-RIP-RST - RF-PART-CONF & 1.31E-10 & 1 & 54 & 8.86E-10 & Rejected \\
                     & RF-RIP-RST - RF-C45-RST & 1.63E-10 & 0 & 54 & 8.86E-10 & Rejected \\
                     & RF-RIP-RST - RF-RIP-NONE & 2.68E-10 & 1 & 52 & 8.86E-10 & Rejected \\
                     & RF-RIP-RST - RF-RIP-CONF & 3.18E-10 & 2 & 51 & 8.86E-10 & Rejected \\
                     & RF-RIP-RST - RF-PART-RST & 2.7E-09 & 4 & 51 & 2.7E-09 & Rejected \\
\thickhline
\end{tabular}
}
\end{table}

\begin{table}[!ht]
\centering
\caption{Friedman $p$-values for all ratios when comparing SlGb (RF-PART-RST) with state-of-the-art semi-supervised classifiers in terms of prediction rates (kappa). There are significant differences for all ratios, where SlGb exhibits the highest mean (see Table \ref{table:sota}).}
\label{table:friedman4}
\begin{tabular}{|c|c|c|}
\thickhline
Ratio & Friedman $p$-value & $H_0$ \\ \thickhline
10\% & 1.91E-06 & Rejected \\
20\% & 7.62E-07 & Rejected \\
30\% & 3.50E-06 & Rejected \\
40\% & 2.19E-03 & Rejected \\ \thickhline
\end{tabular}
\end{table}

\begin{table}[!ht]
\centering
\caption{Wilcoxon $p$-values and Holm's post-hoc correction using SlGb approach as control method against state-of-the-art semi-supervised classifiers. SlGb significantly outperforms other methods except for CT(SMO) and DCT when using 30\% and 40\% of labeled instances.}
\label{table:wilcoxon4}
\resizebox{12cm}{!} {
\begin{tabular}{|c|c|cccc|c|}
\thickhline
Labeled ratio & SSC algorithm & Wilcoxon $p$-value & $R^-$ & $R^+$ & Holm & $H_0$ \\
\thickhline
\multirow{4}{*}{10\%} & CB(C45) & 3.6E-06 & 12 & 43 & 1.44E-05 & Rejected \\
                     & TT(C45) & 4.23E-06 & 13 & 42 & 1.44E-05 & Rejected \\
                     & DCT & 1.86E-05 & 12 & 43 & 3.71E-05 & Rejected \\
                     & CT(SMO) & 1.74E-04 & 16 & 39 & 1.74E-04 & Rejected \\
\thickhline
\multirow{4}{*}{20\%} & CB(C45) & 1.3E-07 & 9 & 46 & 5.21E-07 & Rejected \\
                     & TT(C45) & 8.37E-07 & 9 & 46 & 2.51E-06 & Rejected \\
                     & DCT & 2.77E-04 & 15 & 40 & 5.53E-04 & Rejected \\
                     & CT(SMO) & 2.35E-03 & 19 & 36 & 2.35E-03 & Rejected \\   
\thickhline
\multirow{4}{*}{30\%} & CB(C45) & 1.64E-07 & 9 & 46 & 6.54E-07 & Rejected \\
                     & TT(C45) & 4.84E-07 & 9 & 45 & 1.45E-06 & Rejected \\
                     & DCT & 7.16E-03 & 19 & 36 & 1.43E-02 & Rejected \\
                     & CT(SMO) & 5.4E-02 & 20 & 35 & 5.4E-02 & Not Rejected \\      
\thickhline
\multirow{4}{*}{40\%}  & TT(C45) & 6.58E-05 & 12 & 42 & 2.63E-04 & Rejected \\
                     & CB(C45) & 8.18E-05 & 15 & 39 & 2.63E-04 & Rejected \\
                     & DCT & 0.172 & 24 & 31 & 0.344 & Not Rejected \\
                     & CT(SMO) & 0.633 & 27 & 28 & 0.633 & Not Rejected \\
\thickhline
\end{tabular}
}
\end{table}

\clearpage

\end{document}